\documentclass[lettersize,journal]{IEEEtran}
\usepackage{amsmath,amsfonts}
\usepackage{algorithm}
\usepackage{algorithmicx}
\usepackage{array}
\usepackage{textcomp}
\usepackage{stfloats}
\usepackage{url}
\usepackage{verbatim}
\usepackage{graphicx}
\usepackage{cite}
\usepackage{balance}
\usepackage{caption}
\usepackage{booktabs}
\usepackage{multirow}
\usepackage{graphicx}
\usepackage{adjustbox}
\usepackage{algpseudocode}
\usepackage{amssymb}
\usepackage{amsmath}
\usepackage{fontawesome5}
\usepackage{xcolor}
\usepackage{subfigure}
\usepackage{comment}
\usepackage[numbers,sort&compress]{natbib}
\usepackage[hidelinks,colorlinks=false,breaklinks=true,pagebackref=false,allcolors=black]{hyperref}

\usepackage{bibentry}
\begin{document}

\title{Agent-Aware Training for Agent-Agnostic Action Advising in Deep Reinforcement Learning}

\author{Yaoquan~Wei,
        Shunyu~Liu$^*$,
        Jie~Song,
        Tongya Zheng,
        Kaixuan Chen,
        Yong Wang,\\
        Mingli~Song,~\IEEEmembership{Senior Member, IEEE}
\thanks{$^*$Corresponding author.}
\thanks{Y.~Wei, J.~Song are with the College of Software Technology, Zhejiang University, Hangzhou 310027, China (e-mail: yaoquanwei@zju.edu.cn, sjie@zju.edu.cn).}
\thanks{S.~Liu, K.~Chen, and M.~Song are with the College of Computer Science and Technology, Zhejiang University, Hangzhou 310027, China (e-mail: liushunyu@zju.edu.cn, chenkx@zju.edu.cn, brooksong@zju.edu.cn).}
\thanks{T.~Zheng is with the Big Graph Center, School of Computer and Computing Science, Hangzhou City University, Hangzhou 310015, China (e-mail: doujiang\_zheng@163.com).}
\thanks{Y.~Wang is with the State Grid Shandong Electric Power Company, Jinan 250001, China (e-mail: wangyong@sd.sgcc.com.cn).}
}

\markboth{\tiny{\qquad\quad This work has been submitted to the IEEE for possible publication. Copyright may be transferred without notice, after which this version may no longer be accessible.}}%
{Wei \MakeLowercase{\textit{et al.}}: Agent-Aware Training for Agent-Agnostic Action Advising in Deep Reinforcement Learning}


\maketitle

\begin{abstract}
Action advising endeavors to leverage supplementary guidance from expert teachers to alleviate the issue of sampling inefficiency in Deep Reinforcement Learning~(DRL). Previous agent-specific action advising methods are hindered by imperfections in the agent itself, while agent-agnostic approaches exhibit limited adaptability to the learning agent. In this study, we propose a novel framework called \textit{Agent-Aware trAining yet Agent-Agnostic Action Advising}~(A7) to strike a balance between the two. The underlying concept of A7 revolves around utilizing the similarity of state features as an indicator for soliciting advice. However, unlike prior methodologies, the measurement of state feature similarity is performed by neither the error-prone learning agent nor the agent-agnostic advisor. Instead, we employ a proxy model to extract state features that are both discriminative~(adaptive to the agent) and generally applicable~(robust to agent noise). Furthermore, we utilize behavior cloning to train a model for reusing advice and introduce an intrinsic reward for the {advised} samples to incentivize the utilization of expert guidance. Experiments are conducted on the GridWorld, LunarLander, and six prominent scenarios from Atari games. The results demonstrate that A7 significantly accelerates the learning process and surpasses existing methods~(both agent-specific and agent-agnostic) by a substantial margin. Our code will be made publicly available. 
\end{abstract}

\begin{IEEEkeywords}
Action advising, contrastive learning, deep reinforcement learning.
\end{IEEEkeywords}

\section{Introduction}
\IEEEPARstart{D}eep Reinforcement Learning~(DRL) has emerged as a well-established paradigm for addressing sequential decision-making tasks~\cite{mnih2013playing,barto2020looking} spanning across diverse  practical domains, including video games~\cite{vinyals2019grandmaster,ye2020mastering}, robotics~\cite{sangiovanni2018deep,andrychowicz2020learning}, auto-driving~\cite{chen2019model,kiran2021deep}, industrial control~\cite{zhou2020smart,yang2020optimal,sharma2021} \textit{etc}. DRL necessitates the agent's acquisition of knowledge through trial and error, enabling them to adapt and enhance their performance by interacting with the environment. However, a formidable challenge within the realm of DRL lies in sampling inefficiency ~\cite{yarats2021improving,ye2022improving}, as the agent must engage in numerous interactions with the environment in order to acquire a promising policy.
\begin{figure}[!t]
    \centering
    \includegraphics[width=0.49\textwidth]{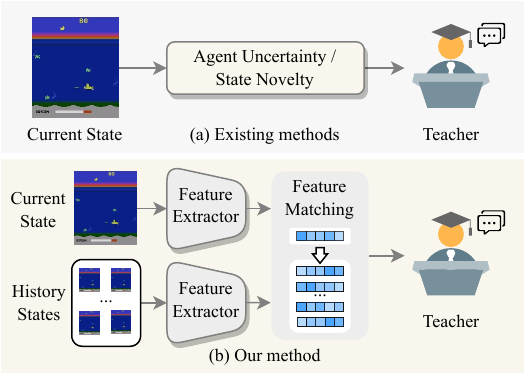}
    \caption{
    Comparing our method with the existing methods for action advising. 
    (a)~Existing methods rely on estimating the uncertainty or novelty for the current state to seek teacher advice. 
    (b)~Our method conducts feature matching to seek teacher advice, which considers the relationship between the current state and history states.
    }
    \label{fig:compare}
\end{figure}

To this date, there has been a remarkable amount of research
effort to overcome the sampling inefficiency with the aid of online expert feedback, including \textit{action-based advice}~\cite{torrey2013teaching,da2020uncertainty,liu2023askac}, \textit{preference-based evaluation}~\cite{christiano2017deep,lee2021pebble,gpt4}, and \textit{language-based instruction}~\cite{goyal2019using,zhou2021inverse}.
Among the diverse approaches, action advising is recently gaining increasing attention as a straightforward yet compelling solution for its more accurate guidance on the policy. Nevertheless, the inherent nature of continuous interactions in action advising inevitably burdens the expert with substantial communications. Hence, the agent must judiciously determine when to seek guidance and effectively leverage the limited resources of expert advice.

Existing action advising approaches determine whether or not acquire action advice from the expert by evaluating the agent uncertainty or  novelty of the current state, as depicted in Figure~\ref{fig:compare}(a), which can be broadly classified into two categories: \textit{agent-specific} methods and \textit{agent-agnostic} methods.
Agent-specific methods~\cite{da2020uncertainty,ilhan2022methodical} hinge upon the agent's inherent uncertainty on the current state to solicit advice from teachers, thus exhibiting enhanced adaptability to the agent policy. Albeit effective in certain scenarios, the uncertainty estimation is easily misled by the agent's own imperfections, consequently resulting in inadequate coverage of the advised state space. 
In contrast, agent-agnostic methods~\cite{ilhan2019teaching,ilhan2021student} assess the novelty of the state from the viewpoint of the demonstrator or others, irrespective of the agent's policy. This circumvents the issues stemming from an imperfect agent~(especially during the early stage of training) yet leads to wastage of advice in states where the agent has already gained~sufficient~experience.

In this work, we endeavor to amalgamate the advantages of both approaches. We present an innovative framework called  \textit{Agent-Aware trAining yet Agent-Agnostic Action Advising}~(A7) for predicting state novelty, as depicted in Figure~\ref{fig:compare}(b).
The fundamental concept of A7 revolves around utilizing the similarity of state features as an indicator for seeking advice. However, unlike previous methodologies, the measurement of state feature similarity is not performed by the error-prone learning agent or the agent-agnostic advisor. We employ a proxy model~(Feature Extractor) to extract state features that are both discriminative~(i.e., adaptive to the agent) and generally applicable~(i.e., robust to agent noise). Taking inspiration from the contrastive method BYOL~\cite{grill2020bootstrap}, we tailor a contrastive learning approach called action-BYOL to train the proxy feature extractor by contrasting the current state with the subsequent state following the agent's policy~(in which sense we call it \textit{agent-aware training}). Upon encountering a new state, action-BYOL extracts its features and conducts feature matching~(compares them with those from historical states), based on which an advice query is sent to an external expert~(in which sense we call it \textit{agent-agnostic action advising}). Additionally, we employ behavior cloning to train a model for reusing advice and introduce an intrinsic reward for the {advised} samples to incentivize the exploitation~of~expert~guidance.

To summarize, the proposed A7 offers several advantages over state-of-the-art approaches in action advising:
\begin{itemize}
    \item In contrast to prior agent-specific methods such as RCMP~\cite{da2020uncertainty} and SUA-AIR~\cite{ilhan2022methodical}, A7 employs a self-supervised learning strategy to acquire generally applicable state features. This agent-agnostic approach reduces sensitivity to imperfections in the learning agent.
    \item In comparison to prior agent-agnostic methods like SNA~\cite{ilhan2019teaching} and ANA~\cite{ilhan2021student}, A7 leverages the proposed action-BYOL to extract state features. This method trains the feature extractor by contrasting the current state with the next state following the agent's policy, resulting in more discriminative features for identifying novel states.
\end{itemize}

Experiments are conducted on the GridWorld, LunarLander, and six popular scenarios from Atari games~(including Enduro, Freeway, Pong, Qbert, Seaquest, and SpaceInvaders). The results demonstrate that the proposed A7 framework significantly accelerates the learning process and surpasses existing~(both agent-specific and agent-agnostic) methods by a substantial margin.

\section{Related Work}

We briefly review recent advances that are most related to this work, including reinforcement learning from human feedback and contrastive learning.

\subsection{Reinforcement Learning from Human Feedback}
To overcome the sampling inefficiency problem in DRL, learning from human feedback has attracted much attention in the academic field in recent years, where human feedback can be roughly divided into \emph{action-based advice}~\cite{arora2021survey,liu2023askac}, \emph{preference-based evaluation}~~\cite{christiano2017deep,gpt4}, and \emph{language-based instruction}~~\cite{goyal2019using,zhou2021inverse}.
~\citeauthor{christiano2017deep} first scaled preference-based learning to utilize modern deep learning techniques while \citeauthor{lee2021b} proposed a feedback-efficient RL algorithm by utilizing off-policy learning and pre-training for preference-based methods recently.
Toro Icarte et al.~\cite{toro2018advice} utilize natural language advice~(e.g., ``Turn out the lights before you leave the office” or ``Always alleviate potholes in the road”), which can recommend regarding behavior to guide the exploration~of~the~RL~agent.

Compared with the low discriminability of preference-based evaluation and the semantic ambiguity of language-based instruction, we are interested in action-based advice methods, also called action advising, which provides much more accurate guidance to the agent.
At the heart of action-advising methods is how to determine the optimal timing for the student agent to solicit action advice from the teacher model~(a pre-trained model or an expert).
\emph{Agent-specific} and \emph{agent-agnostic} methods have dominated the two mainstream branches of action advising, which both assess the advice significance based on the uncertainty of the current state.
On the one hand, agent-specific methods evaluate the agent-level uncertainty based on the current state from the agent network.
\citet{torrey2013teaching} initially estimated the uncertainty of the teacher agent by considering its Q-value and sought advice upon high uncertainty. In contrast, \citet{da2020uncertainty} calculated the uncertainty from the viewpoint of the student agent based on an multi-head attention network employed by Bootstrapped DQN~\cite{osband2016deep}.
\citet{liu2023askac} additionally employed the value loss as a measure of state uncertainty.
\citet{ilhan2022methodical} further calculated the uncertainty by utilizing a twin network with dropout to mitigate interference from the original network.

On the other hand, agent-agnostic methods~\cite{ilhan2019teaching,ilhan2021student} evaluate state-level uncertainty based on the global states beyond the limited viewpoint of a single agent. \citet{ilhan2021student} measured the novelty of a piece of advice based on Random Network Distillation~(RND) and only updated RND for the advised states.
In addition, there are some other methods. For example, \citet{torrey2013teaching} introduced a teacher uncertainty method that uses the Q-value of the teacher agent to decide when to get the advice. However, this method requires that the teacher should also be a model, so its application in real life is relatively narrow.
Albeit effective of existing action-advising approaches, we are motivated to bridge the advantages of agent-specific and agent-agnostic methods in this paper to advance the utility of expert feedback.  \citet{torrey2013teaching} introduced a teacher uncertainty method that uses the Q-value of the teacher agent to decide when to get the advice.

Moreover, action advising methods often employ policy shaping to optimize the performance of the agent.
Policy shaping allows for the direct integration of external feedback into the learning process~\cite{bignold2023conceptual}, modifying Q-values or installing rules to supersede specific actions in selected states. Therefore, one advantage of the policy shaping approach is that it does not rely on the representation of the problem using a reward function. In some real-life scenarios with many conflicting objectives, the policy shaping approach can make it easier for the agent to indicate if its policy is correct directly, rather than trying to explain it through some implicit reward functions.
\citet{griffith2013policy} inferred the optimal policy based on the binary labels~(right or wrong) from human feedback; 
\citet{harnack2022quantifying} used human feedback as a mistake correction to guide agent exploration and explored the optimal range of feedback frequencies;
\citet{bignold2021persistent} further allowed humans to provide guidance in the form of rules and even provide advice in advance without needing to consider matching conditions for the state.
Among them, action advising involves directly altering the learned behavior of the~agent~for~exploration. 

\subsection{Contrastive Learning}
The data-thirsty requirement of high-quality annotated labels has posed a significant challenge to existing \emph{state-of-the-art} deep learning models~\cite{he2016deep}.
Recently, contrastive learning~\cite{wu2018unsupervised,oord2018representation,ye2019unsupervised,tian2020contrastive} has shown its effectiveness in deep representation learning as a self-supervised paradigm, which even surpasses the \emph{state-of-the-art} fully-supervised deep models on ImageNet.
The pioneering works of contrastive learning push away the positive samples from the negative samples in the latent representation space, relying on a large number of negative samples to obtain high-quality representations, such as MOCO~\cite{he2020momentum} and SimCLR~\cite{chen2020simple}.
Another intriguing direction to bypass the need for constructing positive and negative samples~\cite{caron2020unsupervised,grill2020bootstrap} has presented its effectiveness in computer vision.
SwAV~\cite{caron2020unsupervised} used cluster centers to act as negative prototypes, and BYOL~\cite{grill2020bootstrap} utilized different views of the same sample to contrast their representations away.
Motivated by the highly dynamic states of DRL, we have tailored an action-BYOL for action advising, which avoids the bias of negative sampling in existing contrastive learning methods and learns rich representions from dynamic states.

\section{Background}

In this work, we focus on the action advising problem for the control tasks with the discrete action space under the Markov Decision Process~(MDP).

\paragraph{Markov Decision Process}
In the context of DRL, the sequential decision-making task can be formulated as a MDP. A standard MDP is represented by the tuple $ (\mathcal{S}, \mathcal{A}, \mathcal{R}, \mathcal{P}, \mathbf{\gamma})$ where $\mathcal{S}$ is the state space, $\mathcal{A}$ is the action space, $\mathcal{R}$ is the reward function, $\mathcal{P}$ is the state transition function, and $\mathbf{\gamma}$ is the discount factor. At each time step $t$, the agent can observe the state $s_t \in \mathcal{S}$ and execute the action $a_t \in \mathcal{A}$ according to its policy $\pi\colon \mathcal{S} \rightarrow \Delta(\mathcal{A})$. $\Delta(\mathcal{A})$ denotes the set of probability distributions over the action space $\mathcal{A}$. Then the agent receives the reward $r_t = \mathcal{R}(s_t,a_t)$ from the environment and transitions to the next state $s_{t+1}\sim \mathcal{P}(s_{t+1} | s_{t}, a_t)$. The goal of the agent is to obtain the optimal policy $\pi^*$ that maximizes the discounted return $\sum_{k=0}^{\infty} \gamma^{k} r_{t + k}$.  

\paragraph{Deep Q Network}
The value-based DRL methods tend to assess the quality of a policy $\pi$ by the action-value function $Q$. By obtaining the optimal action-value function, we can deliver an optimal policy $\pi^*$ for the learning agent. To estimate the optimal action-value function, Deep Q Network~(DQN)~\cite{mnih2013playing} uses a neural network $Q_{\omega}$ with parameters $\omega$ as an approximator. Specifically, the network is optimized by minimizing the following Temporal-Difference~(TD) loss based on the Bellman equation:
\begin{equation}
    \mathcal{L}_Q =  \mathbb{E}_{\mathcal{D}_Q}\left[(r_{t} + \gamma \max_{a'} Q_{\hat{\omega}}(s_{t+1}, a') - Q_{\omega}(s_{t}, a_t))^2\right],
\end{equation}
where $\mathcal{D}_Q$ is the replay buffer and $\hat{\omega}$ is the parameters of the target network periodically updated by the online parameters $\omega$~\cite{mnih2013playing}. 
In this paper, we adopt a DQN variant, dueling DQN~\cite{wang2016dueling}, as the backbone of all compared methods~(except RCMP) to ensure comparability.

\paragraph{Action Advising} In the framework of action advising in DRL, a student agent $\pi_S$ can seek action advice from the expert teacher $\pi_T$ to learn an effective policy. Then the expert teacher will return the action advice $\tilde{a}_t = \pi_{T}(s_t)$ based on the current state $s_t$. Specifically, the advice budget $N$ is often limited due to resource constraints.

\begin{figure*}[!t]
	\includegraphics[width=1\textwidth]{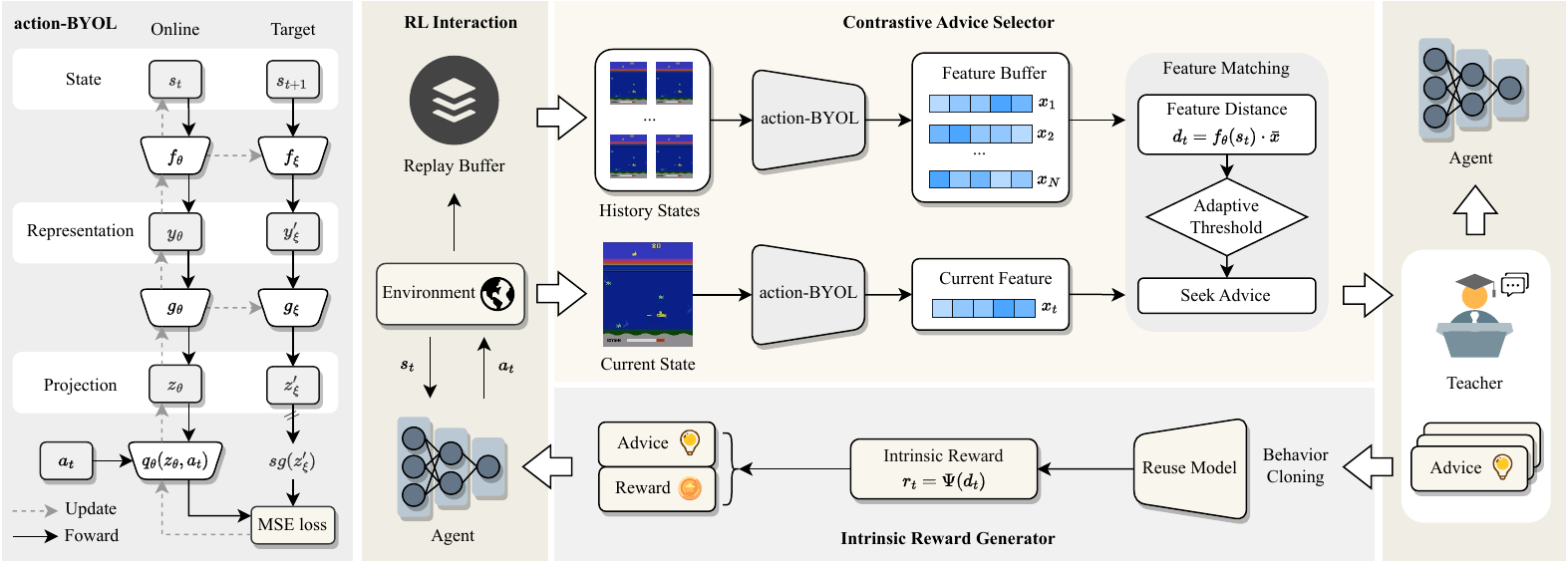} 
	\caption{\textbf{Left:} an illustrative diagram of action-BYOL, which minimizes a similarity loss between $q_\theta(z_\theta,a_t)$ and $sg(z'_\xi)$. $sg(\cdot)$ means stop-gradient operation. \textbf{Right:} an illustrative diagram of the proposed \textit{Agent-Aware trAining yet Agent-Agnostic Action Advising}~(A7) framework, comprising two key components: the contrastive advice selector and the intrinsic reward generator.}
	\label{fig:workflow}
\end{figure*}

\section{Method}
In what follows, we detail the proposed \textit{Agent-Aware trAining yet Agent-Agnostic Action Advising}~(A7) framework. As shown in Figure~\ref{fig:workflow}, A7 comprises two complementary components: the contrastive advice selector and the intrinsic reward generator. 
The contrastive advice selector employs a proxy model called action-BYOL, which is trained with the states experienced by the agent~(agent-aware) to extract relevant state features. 
Then, the selector only uses the similarity between state features to identify an appropriate state for seeking advice, regardless of the agent~(agent-agnostic).
Moreover, the intrinsic reward generator collects the state-advice pairs chosen by the selector for reuse. It also introduces additional intrinsic rewards for advised samples~(advised by the teacher and reuse model) to incentivize the exploitation of expert guidance. With these two components, A7 can accelerate the agent learning process and improve sampling efficiency. The description of our framework is shown in Algorithm \ref{alg:FADD}.

\subsection{Contrastive Advice Selector}
To integrate the benefits of existing methods, encompassing adaptability to agent behavior and robustness to agent noise, we adopt the similarity among state features as an indicator for seeking action advice.
This necessitates the effective extraction of state features in our approach. To achieve this, we employ the contrastive learning method, BYOL~\cite{grill2020bootstrap}, to train the feature extractor using states experienced by the agent, which can be referred to as agent-aware training. However, the similarity calculation for action advising is performed without considering the specific agent, which can be referred to as agent-agnostic action advising. Contrastive learning has gained popularity in learning rich representations of high-dimensional data for downstream tasks. Moreover, considering the temporal relationships between states, we introduce modifications to the BYOL method. When an agent encounters a state and takes an action, it transitions to the next state. {Two consecutive states are usually similar.} In addition, the selected action can provide transitional information between the current state and the subsequent state. To fully leverage this information, we incorporate the current state, selected action, and next state into the contrastive learning process instead of employing simple state augmentations. We term this modified model as action-BYOL. 
{It is also worth noting that the hidden layer of the agent network also has the potential for feature extraction. Nonetheless, the continuous update of network parameters, along with incomplete initial network training, limits its effectiveness in representing the relationship between states. Therefore, we adopt a separate pre-trained action-BYOL as a feature extractor.}

The action-BYOL model consists of the \textit{online} and \textit{target} networks, as depicted in the left part of Figure~\ref{fig:workflow}. 
The online network with parameters $\theta$ takes the current state $s_t$ as input and outputs the representation $x_\theta \triangleq f_\theta(s_t)$, as well as the projection $z_\theta \triangleq g_\theta(x_\theta)$. The target network with parameters $\xi$ takes the next state $s_{t+1}$ as input and outputs the target representation $x'_\xi \triangleq f_\xi(s_{t+1})$, as well as the target projection $z'_\xi \triangleq g_\xi(x'_\xi)$. 
Moreover, we further output a predictor $q_\theta(z_\theta, a_t)$, which takes the selected action $a_t$ and the projection $z_\theta$ as input. Note that the predictor is only applied to the online network. We normalize $q_\theta(z_\theta, a_t)$ and $z'_\xi$ to $\bar{q_{\theta}}(z_\theta, a_t) \triangleq q_\theta(z_\theta, a_t) / \Vert q_\theta(z_\theta, a_t) \Vert_2$ and $\bar{z}'_\xi \triangleq z'_\xi / \Vert z'_\xi \Vert_2$.  Finally, the similarity loss function~\cite{grill2020bootstrap} between the predictions and target projections is defined as:
\begin{equation}
    \mathcal{L}_C =  \Vert \bar{q_{\theta}}(z_\theta, a_t) - \bar{z}'_\xi \Vert_2 ^2 = 1 - \frac{\langle q_\theta(z_\theta, a_t),  z'_\xi \rangle }{\big\|q_\theta(z_\theta, a_t)\big\|_2\cdot
    \big\|z'_\xi\big\|_2
    }
    \cdot
    \label{eq:byol-loss}
\end{equation}
The optimization is performed to minimize $\mathcal{L}_C$ with respect to the online parameters $\theta$ only, while the target parameter $\xi$ is updated slowly by the online parameters via $\xi \leftarrow \tau\xi + (1 - \tau) \theta$ with the target decay rate $\tau \in [0, 1]$.

To facilitate the learning of sample features, we train the action-BYOL model periodically until the budget is exhausted. However, the agent often fails within a few steps during the early stage of training, resulting in the collection of similar samples that hinder the learning process of action-BYOL.
Therefore, to acquire diverse samples, we allow the agent to continuously seek advice from the teacher at the beginning.
After each training stage of action-BYOL, we only retain the encoder $f_\theta$ and use the representation output as the state feature.
The features of all experienced states are stored in a feature buffer $\mathcal{D}_f$. 
At each time step $t$, we extract the agent-agnostic feature from the current state and calculate the average cosine distance between the current feature and the stored features as the indicator for seeking advice:
\begin{equation} 
    d_t = \Phi(s_t, \mathcal{D}_f) = \frac{ \sum_{j=1}^{M} x_j \cdot f_{\theta}(s_{t})}{M},
    \label{equ:dist}
\end{equation}
where $s_t$ is the current state, $x_j \in \mathcal{D}_f$ is the stored feature in the feature buffer, and $M$ is the buffer size. 
{Training an agent involves an extensive process of interactive learning with substantial samples. 
Hence, it is essential to evaluate the overall relationships among the samples. 
}
However, it is not practical to store all state features in a buffer due to the memory overhead and computational speed limitations. Therefore, we transform Equation~(\ref{equ:dist}) into:
\begin{equation}
    d_{t} = f_{\theta}(s_{t}) \cdot \bar{x},\; \text{where} \; \bar{x} = \frac{\sum_{j=1}^{M} x_j}{M} \; \text{and} \;  x_j \in \mathcal{D}_f.
    \label{equ:dist_easy}
\end{equation}
It is easy to devise incremental formulas for updating the average feature $\bar{x}$ with low computational cost. 
Thus, it suffices to store a single average feature $\bar{x}$, eliminating the need to store all individual state features.

During training, the agent will seek the expert teacher for advice if the feature distance $d_t$ of the current state $s_t$ exceeds a threshold $\sigma$. Additionally, the current state feature $x_t=f_\theta({s_t})$ will be used for updating the average feature $\bar{x}$. 
However, it is challenging to determine a fixed distance threshold $\sigma$ for different environments with various feature spaces. Therefore, to address the necessity of tuning the distance threshold for each environment, we employ an adaptive distance threshold. 
At each time step $t$, we add the distance $d_t$ to a fixed-length queue, denoted as $\mathcal{H}$. If the queue is not yet full, the agent can continuously seek advice from the teacher. Once the queue $\mathcal{H}$ reaches its maximum length, we sort the queue $\mathcal{H}$ incrementally and use the percentile value of $\mathcal{H}$ as our adaptive threshold for subsequent steps. Although the length of the queue and the percentile value still require configuration, they can be universally applied across~all~environments.

\subsection{Intrinsic Reward Generator}
To encourage the exploitation of expert guidance, we employ behavior cloning to train a reuse model to imitate the expert teacher for action advising. Moreover, additional intrinsic rewards are introduced for each {advised} sample to train the agent.
Specifically, we collect the state-advice pairs generated by the contrastive advice selector. These pairs are then used to train a neural network known as the reuse model using behavior cloning. Behavior cloning approximates the conditional distributions of actions based on the associated state. 
The reuse model is trained to minimize the negative log-likelihood loss function as:
\begin{equation}
    \mathcal{L}_G=\sum_{(s,a) \in \mathcal{T}}-\log G (a|s; \phi),
\end{equation}
where $\mathcal{T}$ denotes the collected state-advice pairs and $\phi$ represents the parameter of the reuse model $G$. Subsequently, by taking the current state as input, the uncertainty of the reuse model $G$ can be calculated. Then we determine whether to provide its output as re-advice to the student based on the level of uncertainty. A smaller level of uncertainty indicates a high alignment between the current state and the training states of the reuse model, enabling the reuse model to deliver the teacher action of that specific state. 
Specifically, we adopt the dropout layer in the reuse model and utilize $K$ forward passes to calculate the uncertainty of the reuse model under different dropout masks:
\begin{equation}
     \bf{F} = 
    \begin{bmatrix}
    Q_1(s,a_1) \: Q_1(s,a_2) \hdots Q_1(s, a_{|\mathcal{A}|}) \\
    Q_2(s,a_1) \: Q_2(s,a_2) \hdots Q_2(s, a_{|\mathcal{A}|}) \\
    \vdots \\
    Q_{K}(s,a_1) \: Q_{K}(s,a_2) \hdots Q_{K}(s,a_{|\mathcal{A}|})\\
    \end{bmatrix},
\end{equation}
where $|\mathcal{A}|$ denotes the size of the action space and we set $K$ to 100. Then we calculate the variance of each action $a$:
\begin{equation}
     \bf{Q} = 
    \begin{bmatrix}
    var\left(Q(s,a_1)\right) \: var(Q(s,a_2)) \cdots var\left(Q(s, a_{|\mathcal{A}|})\right)
    \end{bmatrix}.
\end{equation}
Thus, reuse model uncertainty $u_s$ on the state $s$ is defined as the average of the variance:
\begin{equation}
    u_s = \frac{\sum_{i=1}^{|\mathcal{A}|} var\left(Q(s, a_i)\right) }{|\mathcal{A}|} .
\end{equation}
The uncertainties of all trained state-advice pairs are computed, and the threshold $u_r$ is set as the lower 90\% of these uncertainties. When using the reuse model, the first step is to calculate the uncertainty $u_s$ for the current state. If the $u_s$ is lower below the threshold $u_r$. The resulting output with a deactivated dropout layer is subsequently utilized as advice for the agent.
In this way, the student can seek advice from the reuse model $G$ when encountering states that are similar to the advised samples.

 \begin{algorithm}[!t]
\caption{The proposed A7 framework}
\label{alg:FADD}
\begin{algorithmic}[1]
    \State {\bfseries Input:}     
    advice budget $N$, student policy $\pi_S$, teacher policy $\pi_{T}$, encoder $f_\theta$ of action-BYOL, replay buffer $\mathcal{D}_{Q}$, feature buffer $\mathcal{D}_f$, distance queue $\mathcal{H}$, reuse model $G$
    \State {\bfseries Output:} 
    trained student policy $\pi_S$
    \While{not terminal}
        \State Get the state $s_t$ from the environment
        \State Calculate the action $a_t$ from the agent policy  $\pi_{S}(s_t)$ 
        \State Proceed to Step~(3) if the advice budget $N=0$
        \textcolor{gray}{\State \# Step~(1) calculate feature similarity}
        \State Extract the state feature $x_t = f_\theta(s_t) $
        \State Calculate the feature distance $d_t = \Phi (x_t, \mathcal{D}_f)$ 
        \State Store the state feature to buffer $ \mathcal{D}_f = \mathcal{D}_f \cup x_t$
        \State Calculate the adaptive threshold $\sigma$ using $\mathcal{H}$
        \State Store the feature distance to queue $\mathcal{H} = \mathcal{H} \cup d_t$ 
        \textcolor{gray}{\State \# Step~(2) determine whether seek teacher advice}
        \If{$d_t > \sigma$} 
            \State Replace the action $a_t$ with teacher advice $\pi_{T}(s_t)$
            \State Update the advice budget $N = N - 1$
        \textcolor{gray}{\Statex \quad \: \# Step~(3) determine whether seek reuse advice}
        \ElsIf{reuse model $G$ is highly confident}
        \State Replace the action $a_t$ with reuse advice $G(s_t)$
        \State Calculate the intrinsic reward $\hat{r}_t = \Psi(d_t)$
        \EndIf 
        \State \textcolor{gray}{\# Step~(4) interaction and training}
        \State Execute $a_t$ and obtain $r_t$, $s_{t+1}$ from the environment    
        \State Store the transition~($s_t$, $s_{t+1}$, $a_t$, $r_t+\hat{r}_t$) to $\mathcal{D}_{Q}$
        \State Update the agent policy $\pi_{S}$ using batch from $\mathcal{D}_Q$
    \EndWhile
\end{algorithmic}
\end{algorithm}

Although the agent can directly execute the re-advised actions from the reuse model to the environment for guidance, 
the standard rewards from the environment are not sufficient for the agent to learn these expert behaviors effectively.
To further encourage the exploitation of these re-advised samples,  we propose to assign intrinsic rewards to each re-advised sample based on its distance from the feature buffer. Specifically, it is necessary for the agent to learn from hard samples. This implies that samples with greater feature distance require a larger intrinsic reward. Thus, the additional intrinsic reward is designed as follows:
\begin{equation}
    \hat{r}_t = \Psi(d_t) = \lambda_t \cdot \tanh{(\frac{d_t}{d_m})},
    \label{equ:intri_rewards}
\end{equation}
where $d_t$ denotes the feature distance between the current state and the stored features. $d_m$ denotes the average feature distance in the feature buffer, serving as a regularization term.
The time decay coefficient $\lambda_t$ controls the effect of intrinsic rewards and decays over time. The speed of decaying $\lambda_t$ determines how long the intrinsic rewards will continue to influence the agent policy. Choosing the appropriate decay speed of $\lambda_t$ can accelerate learning while preventing substantial biases in the policy. In this paper, we adopt a linear decay regime to gradually reduce the value of $\lambda_t$ from the initial value $\lambda_0$. {For advice directly obtained from the teacher, we keep this initial value unchanged.}
By incorporating advice reuse and leveraging intrinsic rewards, the intrinsic reward generator can enhance the effective utilization of teacher advice and expedite the learning process.

\section{Experiments}
To demonstrate the effectiveness of the proposed A7 framework for action advising in DRL, we conduct experiments on the GridWorld~\cite{ilhan2021student}, LunarLander~\cite{towers_gymnasium_2023} and six popular scenarios from Atari games~(\emph{i.e.} Enduro, Freeway, Pong, Qbert, Seaquest and SpaceInvaders) in line with the previous works~\cite{da2020uncertainty,ilhan2021student,ilhan2021action,ilhan2022methodical}. 
In this section, we first introduce the compared methods and the special hyperparameter settings. Then the comparison results are reported and analyzed. Moreover, ablation studies are conducted to investigate the advantages of our A7.

\subsection{Experimental Settings}
We compare A7 with various baselines, including:
\begin{enumerate}[]
\item ~\textbf{Heuristic methods}: \textit{No Advising~(NA)}, where the student agent follows its own policy without advice;
\textit{Early Advising~(EA)}, where the student agent always requests advice until the advice budget is exhausted;\textit{Random Advising~(RA)}, where the student agent requests advice with a probability of $50\%$ at every step.
\item ~\textbf{Agent-specific methods}: \textit{Importance-base Action Advising~(IAA)}~\cite{torrey2013teaching}, where the student agent uses the difference between the maximum and minimum values of the Q-values to calculate uncertainty and requests advice based on a predefined threshold;
\textit{Requesting Confident Moderated Policy Advice~(RCMP)}~\cite{da2020uncertainty},
where the student agent uses multi-head DQN to calculate uncertainty and requests advice based on a predefined threshold;
\textit{Student Uncertainty-driven Advising with Advice Imitation \& Reuse~(SUA-AIR)}~\cite{ilhan2021action,ilhan2022methodical}, where the student agent requests advice based on an adaptive uncertainty estimation, paired with an imitation model that is using uncertainty thresholds for advice reuse.
\item ~\textbf{Agent-agnostic methods}: \textit{Advice Novelty-Based Advising~(ANA)}~\cite{ilhan2021student}, where the student agent adopts random network distillation~\cite{burda2018exploration} to calculate state novelty for action advising. 

\end{enumerate}

The detailed hyperparameters are given in Table~\ref{tab:grid}, where the common training parameters across different methods are consistent to ensure comparability. We adopt Double DQN~\cite{van2016deep} as the basic algorithm backbone.
All compared methods, except for RCMP with a multi-head output layer, share the same network architecture, including three convolutional layers, a fully-connected hidden layer, and a dueling output layer. 
To carry out sufficient experiments, we follow the same teacher setting as previous works~\cite{da2020uncertainty,ilhan2022methodical} to use a pre-trained model as a teacher.  
The advice budget for GridWorld, LunarLander, and Atari games is set to 5k, 5k, and 25k, respectively. These teachers obtain average evaluation scores of 1.0 for GridWorld, 275 for LunarLander, 1556 for Enduro, 28.8 for Freeway, 12 for Pong, 3705 for Qbert, 8178 for Seaquest and 959 for SpaceInvaders. 

\begin{table}[!t]
    \centering
    \resizebox{0.49\textwidth}{!}{%
    \begin{tabular}{c|cc}
     \toprule
Hyperparameter                  & {\begin{tabular}[c]{@{}c@{}}GridWorld \\ \& LunarLand\end{tabular}}  & Atari     \\ \midrule
Learning Rate                   & 0.0001 & 0.0000625  \\ \specialrule{0em}{0.5pt}{0.5pt}
Minibatch Size                  & 32      & 32     \\ \specialrule{0em}{0.5pt}{0.5pt}
Discount Factor $\gamma$          & 0.99   & 0.99     \\ \specialrule{0em}{0.5pt}{0.5pt}
Replay Buffer min. Size         & 500     & 10k    \\ \specialrule{0em}{0.5pt}{0.5pt}
Replay Buffer max. Size         & 5k      & 1000k  \\ \specialrule{0em}{0.5pt}{0.5pt}
Target Network Update Frequency & 100     & 7.5k   \\ \specialrule{0em}{0.5pt}{0.5pt}
Initial Epsilon                   & 1.0 & 1.0  \\ \specialrule{0em}{0.5pt}{0.5pt}
Final Epsilon                     & 0.01  & 0.01 \\ \specialrule{0em}{0.5pt}{0.5pt}
Annealed Epsilon Steps                   & 5k & 250k \\ \specialrule{0em}{0.5pt}{0.5pt}
Advice Budget $N$                  & 5k & 25k \\ \bottomrule
    \end{tabular}
    }
    \caption{Hyperparameters in different environments}
    \label{tab:grid}
\end{table}

In action-BYOL, we encode the action as a one-hot vector with the same length as the action space. We train action-BYOL for 20 epochs each time. For GridWorld and LunarLander, we solely use the MLP network to extract features and train action-BYOL every 1k steps until the budget is depleted. For Atari games, the CNN network is employed to extract features and train action-BYOL every 10k steps until the budget is depleted.
In addition, we employ the projection of the current state to predict the next state, distinguishing it from original BYOL~\cite{grill2020bootstrap}. 
Consequently, our loss function will not be bi-directional anymore. 
Besides, all other parameters and details remain consistent with BYOL~\cite{grill2020bootstrap}. Following the training process, we employ action-BYOL to extract features from all trained samples. We apply regularization to the extracted features and store them in the feature buffer. In fact, we simulate the feature buffer using only one vector.
Subsequently, the calculated distances are stored in a fixed-length queue denoted as 
$\mathcal{H}$ at each step. The length of $\mathcal{H}$ is set to 200, and the adaptive threshold $\sigma$ is defined as the 70-th percentile value of the queue.

The reuse model is supervised and trained with state-advice pair data.	Additionally, a dropout layer is incorporated into the model. Initially, the reuse model is trained for 50k epochs when the student receives 500 pieces of advice in GridWorld and LunarLander, or 2.5k pieces of advice in Atari scenarios. Subsequently, the model will undergo an additional 20k epochs of training for every 500 pieces of advice in GridWorld and LunarLander or 2.5k pieces of advice in Atari scenarios. The reuse model obtains the hyperparameters of 0.0001 for the learning rate, 32 for the minibatch size, and 0.35 for the dropout rate. In our experiments, a probability of 0.5 is set for reusing advice because excessive reuse can lead the model to converge to local optima prematurely.

To further mitigate excessive bias introduced by intrinsic rewards, a linear decay regime is employed to gradually diminish $\lambda_t$ until it reaches 0. Once $\lambda_t$ reaches 0, intrinsic rewards will cease to impact the learning process. For GridWorld and LunarLander, this process takes 20k steps, while for Atari scenarios, it requires 1M steps. 
For the GridWorld and LunarLander scenarios, the initial value of $\lambda_0$ is 0.1. For the Freeway, Qbert, Seaquest, and SpaceInvaders scenarios, the initial value of $\lambda_0$ is 0.5. Meanwhile, for the Pong and Enduro scenarios, the initial value of $\lambda_0$ is 0.1 and 0.2, respectively.

\begin{figure*}[!t]
    \centering
    \hspace{4.5mm}
    \subfigure{
    \includegraphics[scale=0.36]{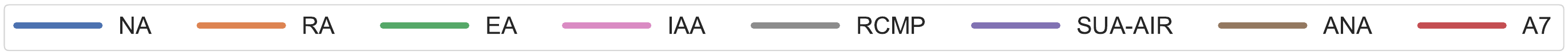}
    \setcounter{subfigure}{0}
    }
    \vspace{-2mm}

    \subfigure[GridWorld]{
    \includegraphics[scale=0.26]{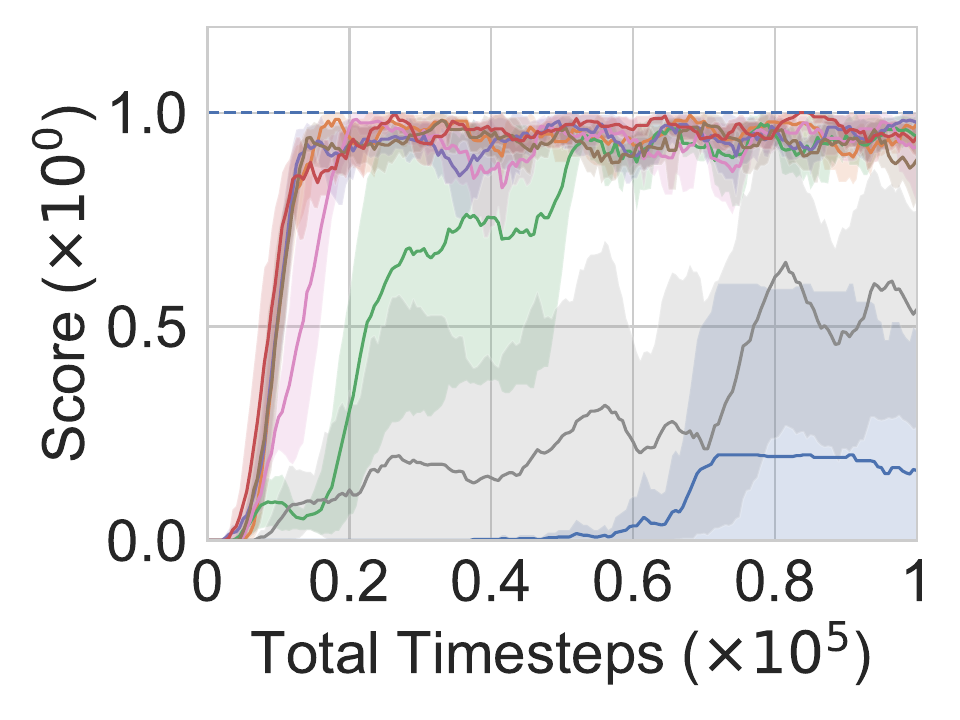}
    }%
    \subfigure[LunarLander]{
    \includegraphics[scale=0.26]{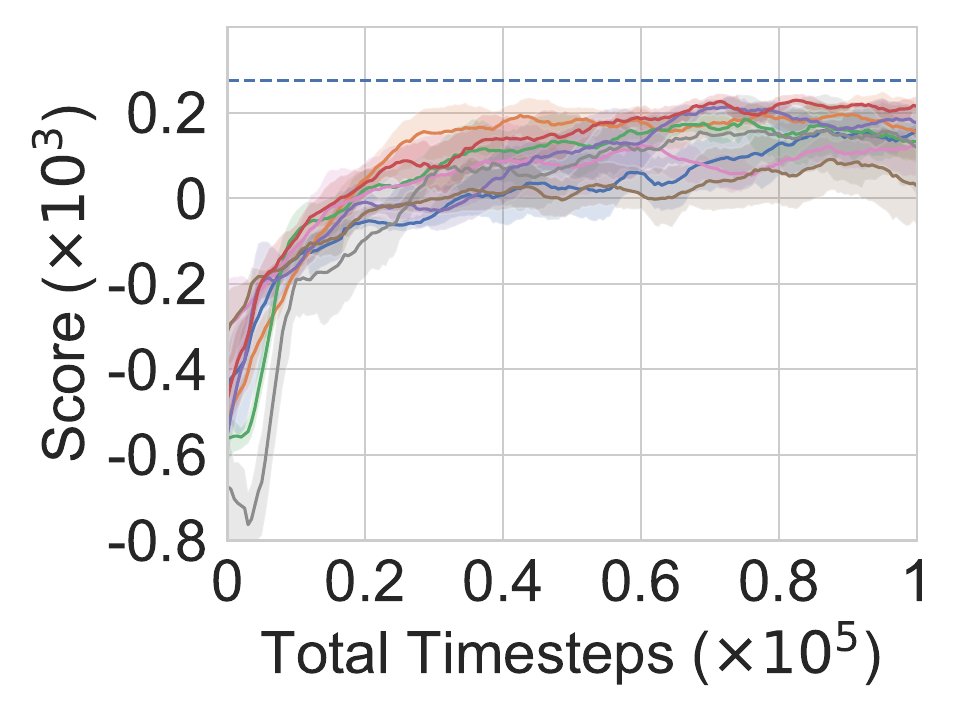}
    }%
    \subfigure[Freeway]{
    \includegraphics[scale=0.26]{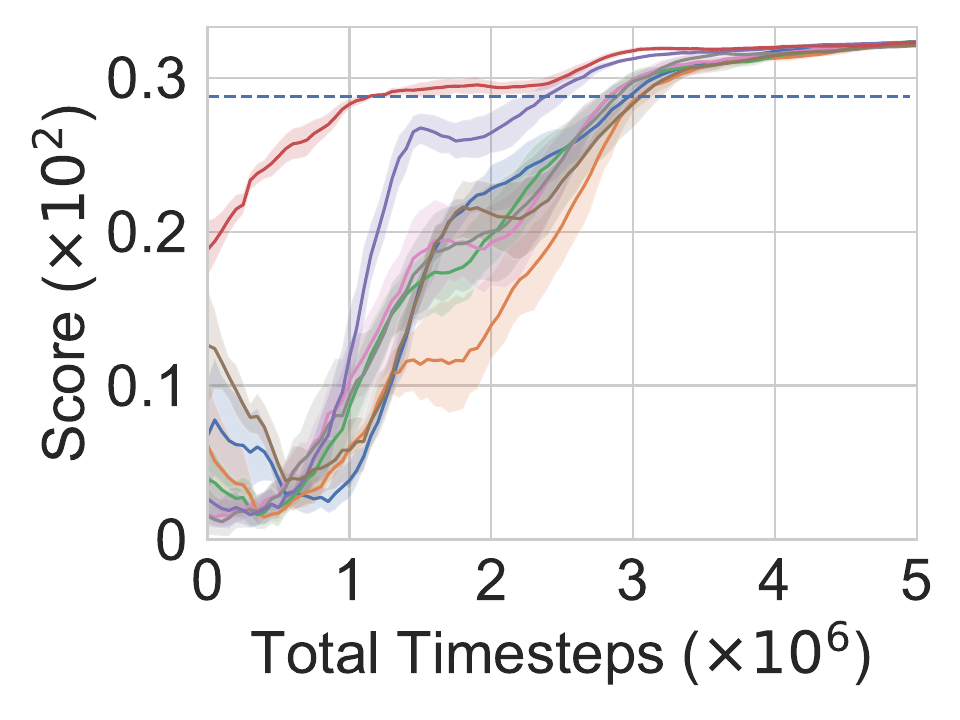}
    }%
    \subfigure[Qbert]{
    \includegraphics[scale=0.26]{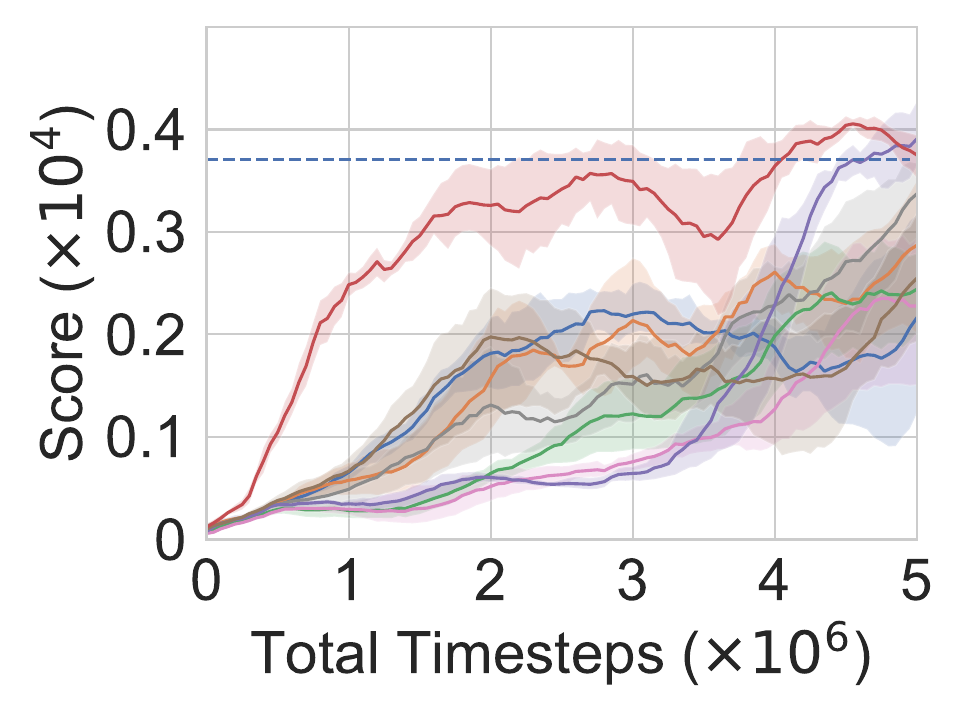}
    }

    \subfigure[Seaquest]{
    \includegraphics[scale=0.26]{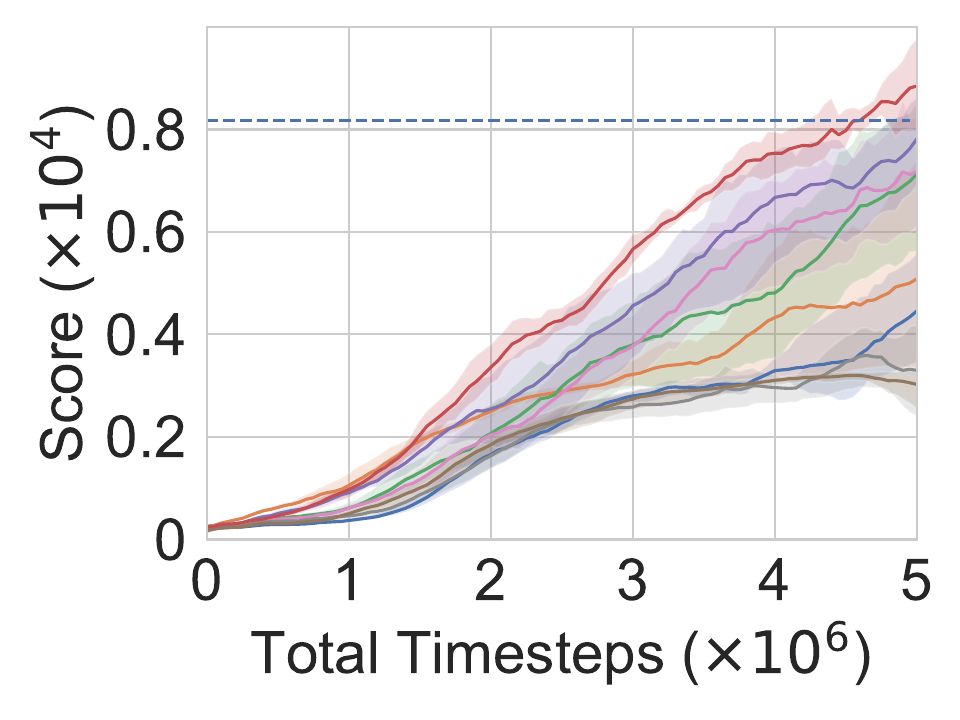}
    }%
    \subfigure[Pong]{
    \includegraphics[scale=0.26]{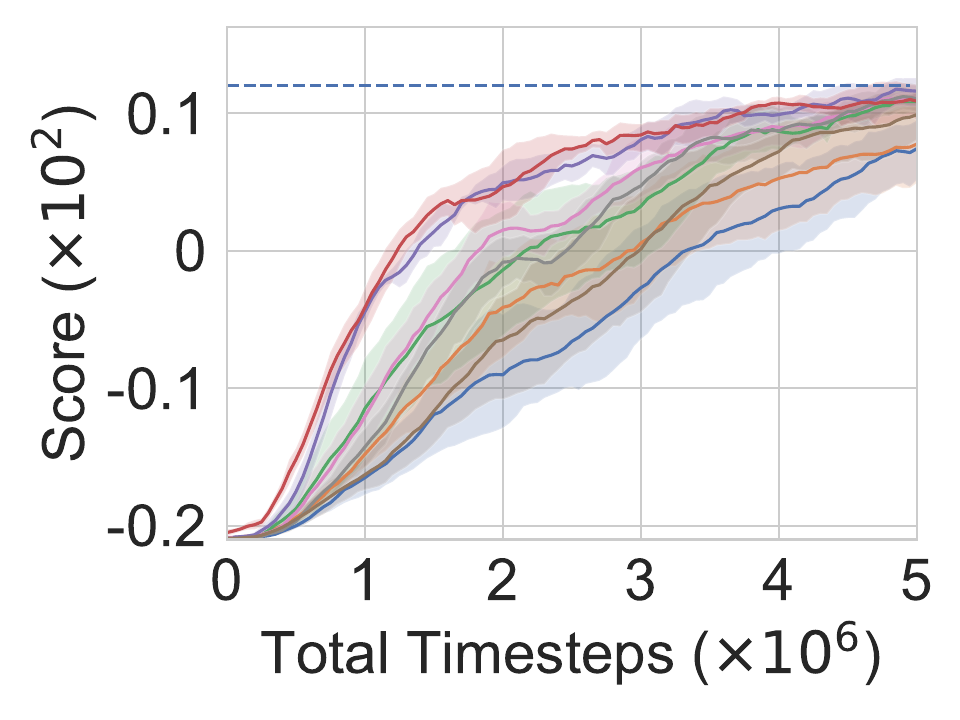}
    }%
    \subfigure[Enduro]{
    \includegraphics[scale=0.26]{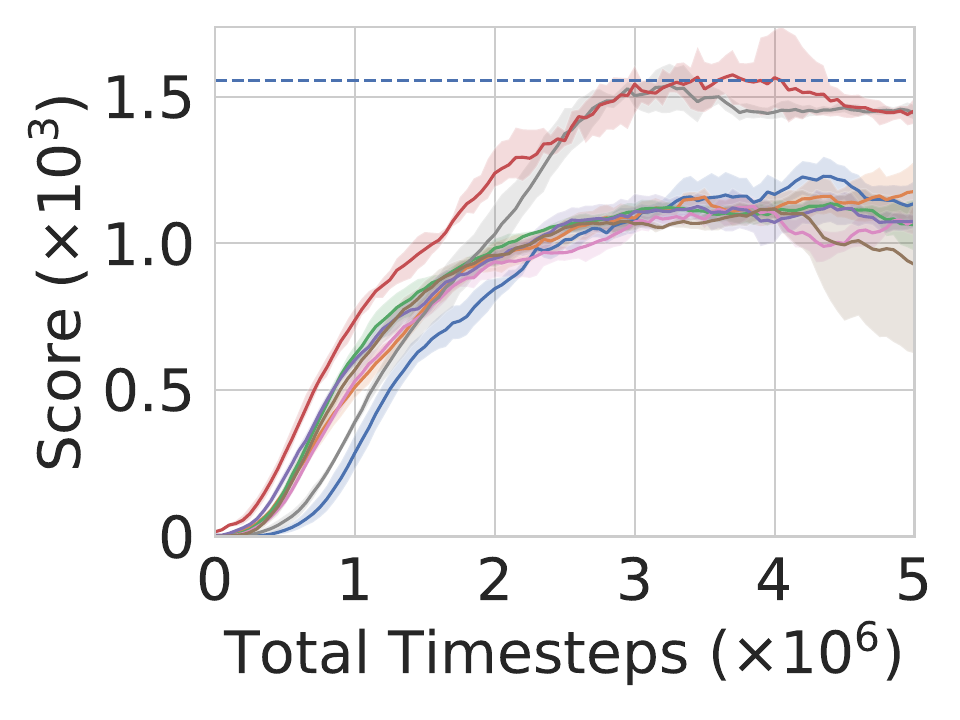}
    }%
    \subfigure[SpaceInvaders]{
    \includegraphics[scale=0.26]{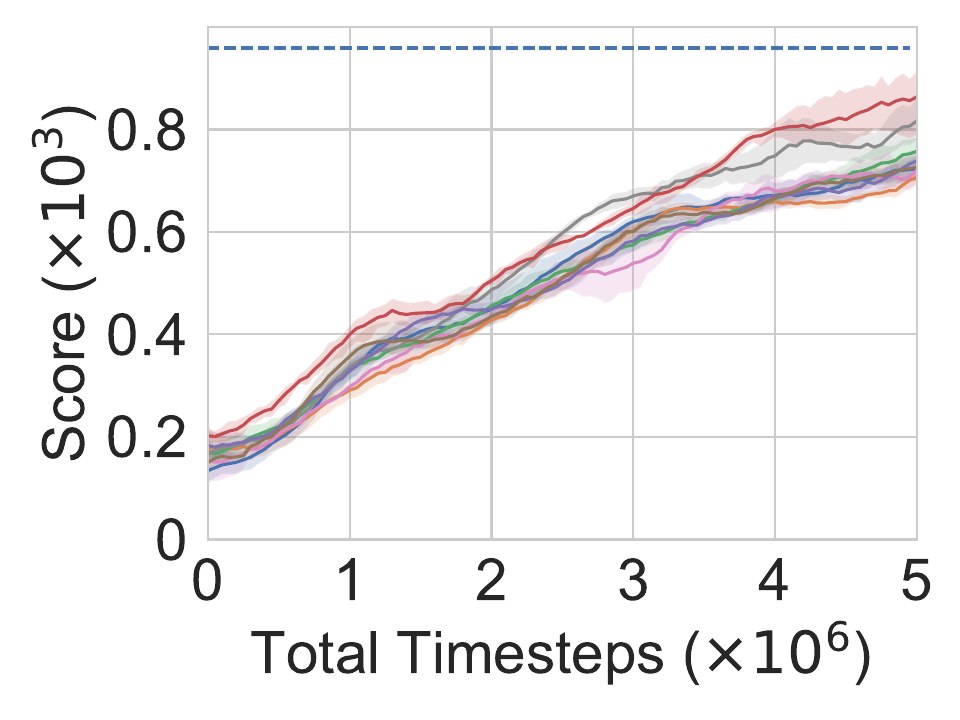}
    }
    \centering
    \caption{Learning curves of our proposed A7 and baselines on the GridWorld, LunarLander, and six Atari scenarios.
    All experimental results are illustrated with the mean and the standard deviation of the performance over five random seeds for a fair comparison. The score represents the cumulative reward for a game during evaluation. To make the results in figures clearer for readers, we adopt a 95\% confidence interval to plot the error region. Dashed lines represent the operation level of the teachers in different environments.}
    \label{fig:result}
\end{figure*}

\begin{table*}[!t]
    \centering
    \resizebox{\textwidth}{!}{%
    \begin{tabular}{@{}ccccccccc@{}}
    \toprule
     \multicolumn{1}{c}{\textbf{Method}} &\textbf{GridWorld} & \textbf{LunarLander} & \textbf{Freeway}  & \textbf{Qbert}    & \textbf{Seaquest}  & \textbf{Pong} & \textbf{Enduro} & \textbf{SpaceInvaders} \\ \midrule
    \textbf{NA}               & 0.07 $\pm$ 0.13 & 0.75 $\pm$ 0.03  & 0.62 $\pm$ 0.01  & 0.27 $\pm$ 0.03  & 0.17 $\pm$ 0.01  & 0.37 $\pm$ 0.07 & 0.45 $\pm$ 0.01 & 0.52 $\pm$ 0.01  \\ \specialrule{0em}{1pt}{1pt}
    \textbf{RA}               & 0.68 $\pm$ 0.11 & 0.83 $\pm$ 0.03   & 0.56 $\pm$ 0.03  & 0.28 $\pm$ 0.04  & 0.23 $\pm$ 0.04  & 0.43 $\pm$ 0.07 & 0.47 $\pm$ 0.01 & 0.50 $\pm$ 0.01 \\ \specialrule{0em}{1pt}{1pt}
    \textbf{EA}               & 0.85 $\pm$ 0.02 & 0.80 $\pm$ 0.03   & 0.62 $\pm$ 0.03  & 0.20 $\pm$ 0.04  & 0.25 $\pm$ 0.07  & 0.50 $\pm$ 0.07 & 0.48 $\pm$ 0.01 & 0.53 $\pm$ 0.01 \\ \specialrule{0em}{1pt}{1pt}
    \textbf{IAA}              &  0.81 $\pm$ 0.01 & 0.78 $\pm$ 0.04  &  0.62 $\pm$ 0.03   & 0.15 $\pm$ 0.04  & 0.25 $\pm$ 0.07  & 0.52 $\pm$ 0.03 & 0.48 $\pm$ 0.01 & 0.50 $\pm$ 0.01 \\ \specialrule{0em}{1pt}{1pt}
    \textbf{RCMP}             &  0.27 $\pm$ 0.22 & 0.75 $\pm$ 0.06  &  0.62 $\pm$ 0.04   & 0.26 $\pm$ 0.06  & 0.25 $\pm$ 0.07  & 0.49 $\pm$ 0.03 &  0.56 $\pm$ 0.01  & 0.57 $\pm$ 0.01\\ \specialrule{0em}{1pt}{1pt}
    \textbf{SUA-AIR}          & 0.85 $\pm$ 0.01  & 0.80 $\pm$ 0.02  & 0.69 $\pm$ 0.01  &  0.22 $\pm$ 0.03 & 0.31 $\pm$ 0.06 & 0.61 $\pm$ 0.02 & 0.48 $\pm$ 0.01 & 0.53 $\pm$ 0.01 \\ \specialrule{0em}{1pt}{1pt}
    \textbf{ANA}              & 0.84 $\pm$ 0.01  & 0.73 $\pm$ 0.04  & 0.62 $\pm$ 0.03  &  0.25 $\pm$ 0.03 & 0.16 $\pm$ 0.01 & 0.43 $\pm$ 0.06 & 0.46 $\pm$ 0.03 & 0.52 $\pm$ 0.01 \\ \specialrule{0em}{1pt}{1pt}
    \midrule
    \textbf{A7}            & \textbf{0.87 $\pm$ 0.02} & \textbf{0.85 $\pm$ 0.01}   & \textbf{0.85 $\pm$ 0.01}  & \textbf{0.54 $\pm$ 0.03}  & \textbf{0.37 $\pm$ 0.01}  & \textbf{0.62 $\pm$ 0.03} & \textbf{0.64 $\pm$ 0.01} &  \textbf{0.60 $\pm$ 0.01}  \\ \specialrule{0em}{1pt}{1pt} 
     \bottomrule

    \end{tabular}
    }
    \caption{ Area under the learning curve~(AUC) of all compared methods in different environments. 
    $\pm$ corresponds to one standard deviation of the average score over five random seeds. \textbf{Bold} indicates the best performance in each environment.}
    \label{tab:auc}
\end{table*}

\begin{table*}[!t]
    \centering
    \resizebox{\textwidth}{!}{%
    \begin{tabular}{@{}ccccccccc@{}}
    \toprule
     \multicolumn{1}{c}{\textbf{Method}} &\textbf{GridWorld} & \textbf{LunarLander} & \textbf{Freeway}  & \textbf{Qbert}    & \textbf{Seaquest}  & \textbf{Pong} & \textbf{Enduro} & \textbf{SpaceInvaders} \\ \midrule
    \textbf{NA}               & 0.18 $\pm$ 0.35                           & 168.52 $\pm$ 25.65 & 32.20 $\pm$ 0.11                           & 1992.40 $\pm$ 1170.26                          & 4461.56 $\pm$ 1234.05                           & 7.61 $\pm$ 2.33                            & 1135.19 $\pm$ \;\,94.63                            & 727.40 $\pm$ 69.29                           \\ \specialrule{0em}{1pt}{1pt}
\textbf{RA}               & 0.91 $\pm$ 0.01                           & 198.45 $\pm$ 35.37 & 32.25 $\pm$ 0.08                           & 2930.50 $\pm$ \;\,702.73                           & 5145.38 $\pm$ 2514.32                           & 7.75 $\pm$ 3.64                            & 1179.58 $\pm$ \;\,86.84                            & 687.05 $\pm$ 35.77                           \\ \specialrule{0em}{1pt}{1pt}
\textbf{EA}               & 0.89 $\pm$ 0.03                           & 185.00 $\pm$ 41.75    & 32.26 $\pm$ 0.15                           & 2563.35 $\pm$ \;\,491.27                           & 6695.36 $\pm$ 1601.95                           & 11.41 $\pm$ 1.04\;\,                           & 1066.53 $\pm$ 105.43                           & 750.70 $\pm$ 53.33                           \\ \specialrule{0em}{1pt}{1pt}
\textbf{IAA}              & 0.86 $\pm$ 0.05                           & 105.00 $\pm$ 13.26  & 31.41 $\pm$ 0.21                           & 2015.00 $\pm$ \;\,573.78                            & 6760.80 $\pm$ 1264.99                            & 4.64 $\pm$ 2.85                            & 1062.61 $\pm$ 168.85                           & 715.50 $\pm$ 35.75                           \\ \specialrule{0em}{1pt}{1pt}
\textbf{RCMP}             & 0.61 $\pm$ 0.02                           & 182.12 $\pm$ 19.41 & 32.14 $\pm$ 0.25                           & 3233.00 $\pm$ 1079.40                            & 2150.80 $\pm$ 1166.01                            & 11.48 $\pm$ 2.03\;\,                           & 1504.75 $\pm$ 268.45 & 875.35 $\pm$ 56.83                           \\ \specialrule{0em}{1pt}{1pt}
\textbf{SUA-AIR}          & 0.94 $\pm$ 0.01                           & 224.95 $\pm$ 63.25 & 32.31 $\pm$ 0.12                           & 4024.95 $\pm$ \;\,606.63 & 7865.12 $\pm$ 1536.11                           & \textbf{12.19 $\pm$ 1.56\;\,} & 1077.28 $\pm$ \;\,92.76                            & 772.95 $\pm$ 76.71                           \\ \specialrule{0em}{1pt}{1pt}
\textbf{ANA}            & 0.91 $\pm$ 0.01                           & \;\;76.32 $\pm$ 15.07  & 32.09 $\pm$ 0.05                           & 2382.35 $\pm$ 1015.09                          & 2950.78 $\pm$ \;\,586.52                            & 9.76 $\pm$ 2.37                            & \;\,920.67 $\pm$ 324.39                            & 730.60 $\pm$ 18.21                           \\ \specialrule{0em}{1pt}{1pt}
\midrule
\textbf{A7}  & \textbf{0.95 $\pm$ 0.01} & \textbf{266.35 $\pm$ 34.12} & \textbf{32.36 $\pm$ 0.15} & \textbf{4096.45 $\pm$ \;\,344.81}                           & \textbf{8692.98 $\pm$ 1009.86} & 11.13 $\pm$ 1.67\;\,                           & \textbf{1544.83 $\pm$ \;\,77.01}                            & \textbf{893.18 $\pm$ 59.24} \\ \specialrule{0em}{1pt}{1pt}
     \bottomrule
    \end{tabular}
    }
    \caption{Test evaluation scores of all compared methods in different environments. 
    $\pm$ corresponds to one standard deviation of the average score over five random seeds. \textbf{Bold} indicates the best performance in each environment.}
    \label{tab:scores}
\end{table*}

\subsection{Results and Analysis}
The experimental results in various environments compared with the state-of-the-art methods are shown in Figure~\ref{fig:result} and Table~\ref{tab:auc}. Specifically, since the curves in Figure~\ref{fig:result} represent the scores of the agents at different time steps, we adopt the area under the learning curve~(AUC) as an important metric to evaluate the sampling efficiency of different methods, which provides an overall measure of the agent's learning efficiency.
In the easy environments~(GridWorld and LunarLander), NA often performs poorly, while our proposed A7 can achieve superior performance. Similarly, several baselines, including EA and SUA-AIR, also exhibit promising results in these two environments.
The exploratory benefit brought by A7 is not obvious. However, in the more difficult environments~(six Atari scenarios), our A7 method offers impressive performance. 
Especially in the Freeway, Qbert, and Seaquest scenarios, our proposed A7 method consistently outperforms baselines by a large margin during training. Moreover, A7 has also demonstrated a substantial performance advantage compared to other methods when evaluated using the AUC metric. In the Freeway and Qbert, our method demonstrates a powerful capability to expedite agent learning, achieving scores that are significantly higher than those obtained by other methods during the early stages of training. In the Seaquest, A7 consistently outperforms other methods in terms of scores throughout the entire duration.
In the Pong, Enduro, and SpaceInvaders scenarios, A7  maintained its leading position in the overall learning curves. Additionally, when considering the overall AUC, A7 remained the best. 
The agent-agnostic method~(ANA) demonstrates good performance in specific environments. However, in other environments like Seaquest and Qbert, its effectiveness is relatively low. This is because it is susceptible to noise interference, making the method less robust.
\begin{figure*}[!t]
    \centering
    \hspace{-6mm}
    \subfigure[GridWorld]{
    \includegraphics[width=0.24\textwidth]{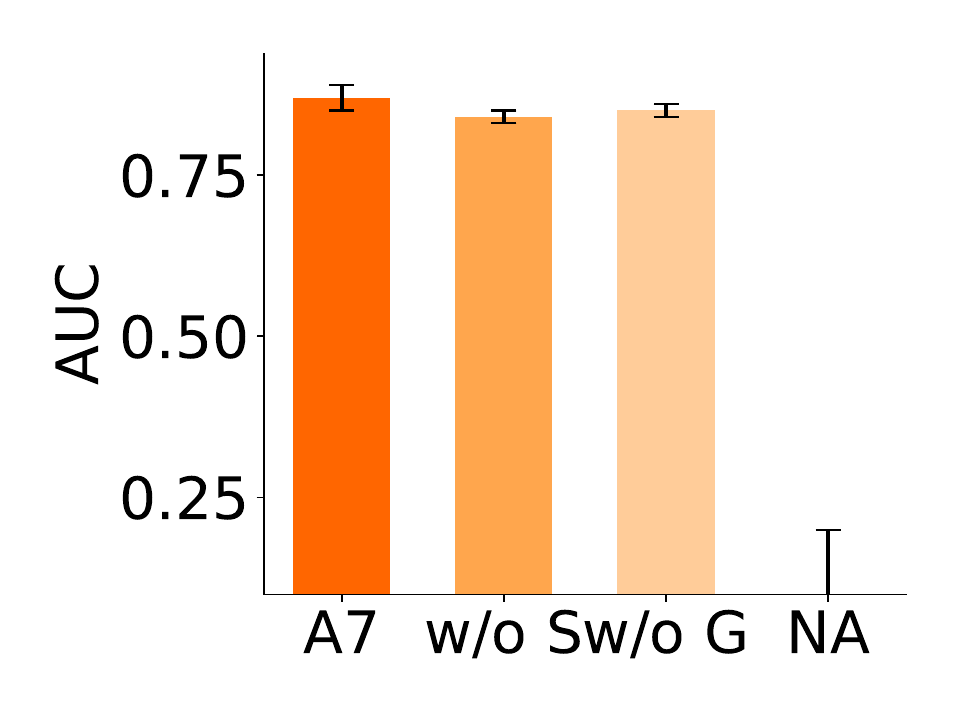}
    }%
    \hspace{-3mm}
    \subfigure[LunarLander]{
    \includegraphics[width=0.24\textwidth]{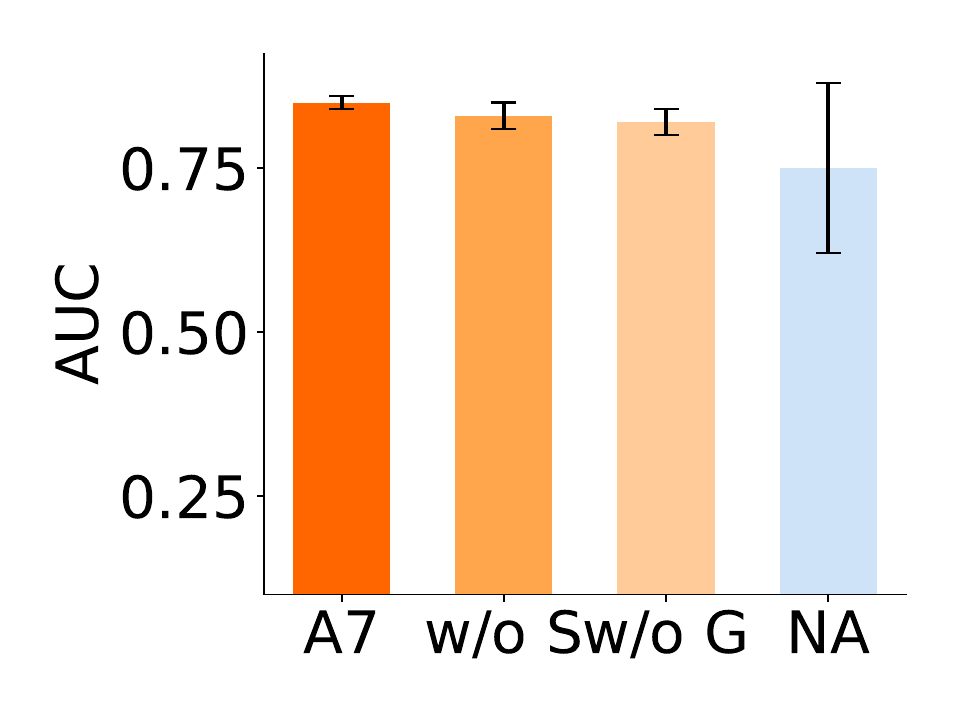}
    }%
    \hspace{-3mm}
    \subfigure[Freeway]{
    \includegraphics[width=0.24\textwidth]{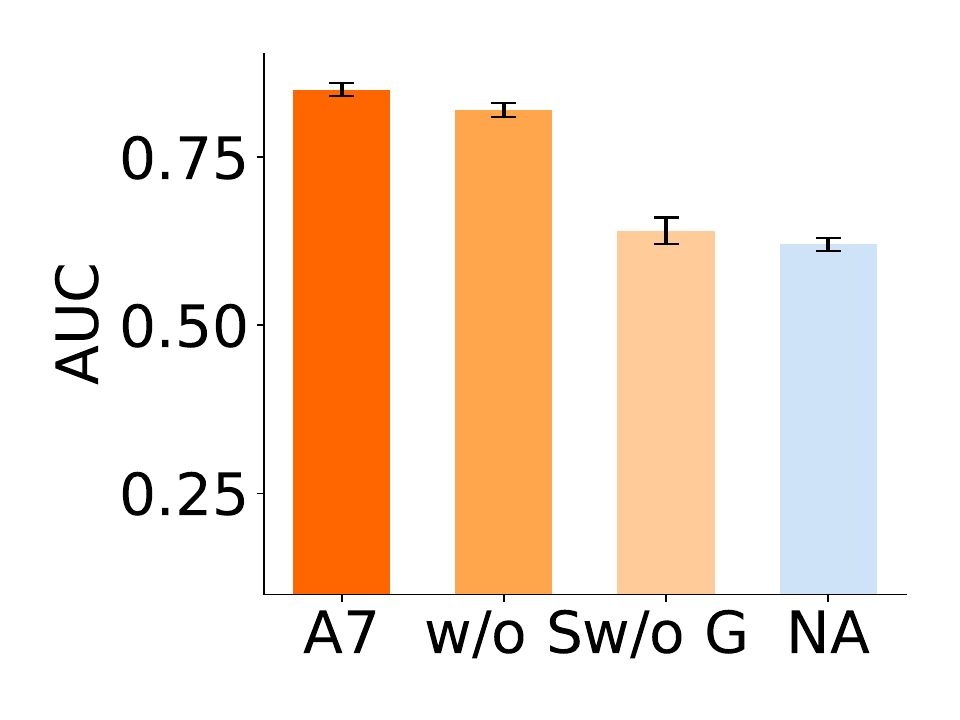}
    }%
    \hspace{-3mm}
    \subfigure[Qbert]{
    \includegraphics[width=0.24\textwidth]{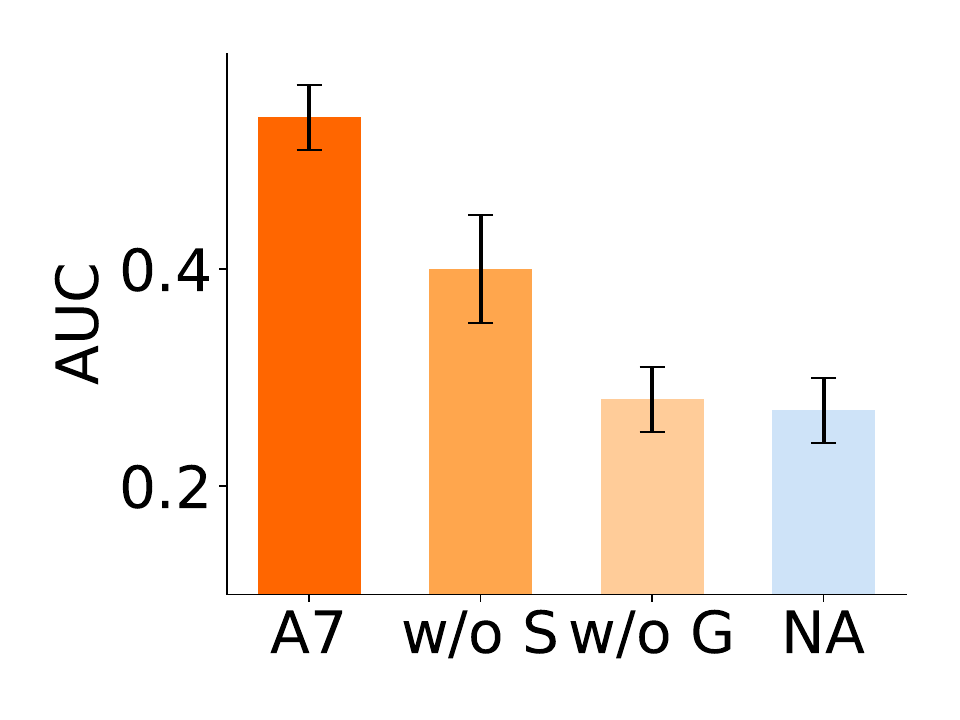}
    }\\
    \hspace{-6mm}
    \subfigure[Seaquest]{
    \includegraphics[width=0.24\textwidth]{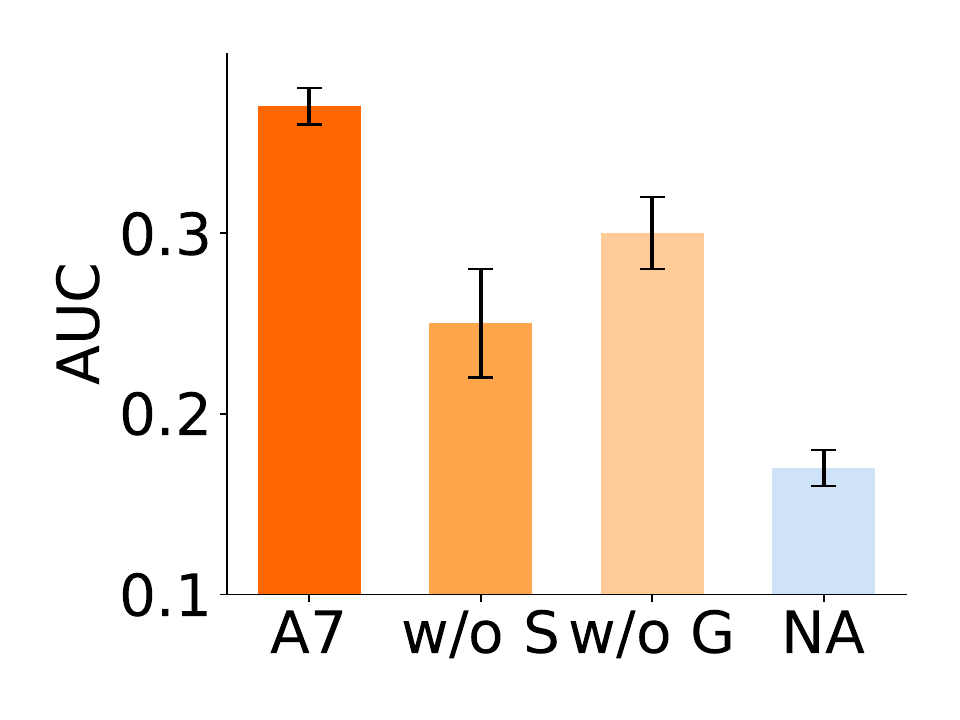}
    }%
    \hspace{-3mm}
    \subfigure[Pong]{
    \includegraphics[width=0.24\textwidth]{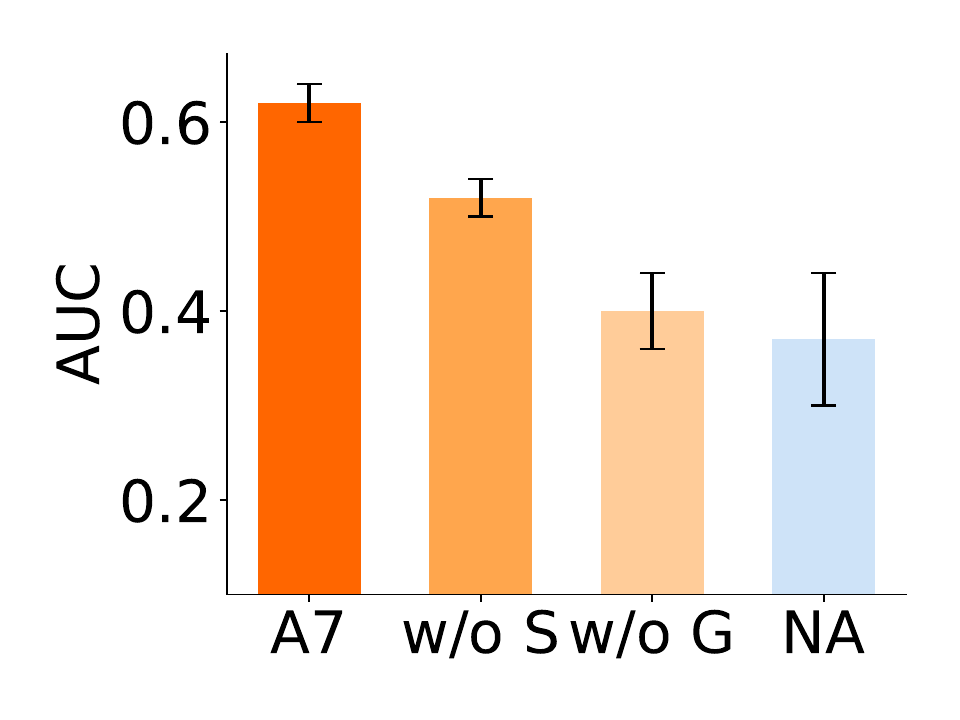}
    }%
    \hspace{-3mm}
    \subfigure[Enduro]{
    \includegraphics[width=0.24\textwidth]{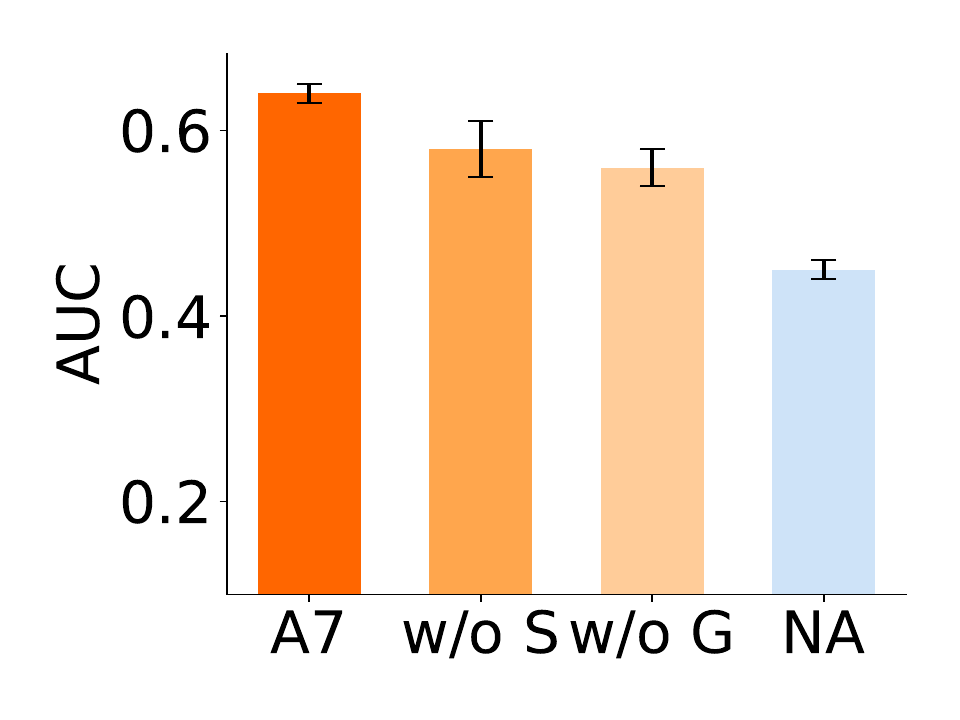}
    }%
    \hspace{-3mm}
    \subfigure[SpaceInvaders]{
    \includegraphics[width=0.24\textwidth]{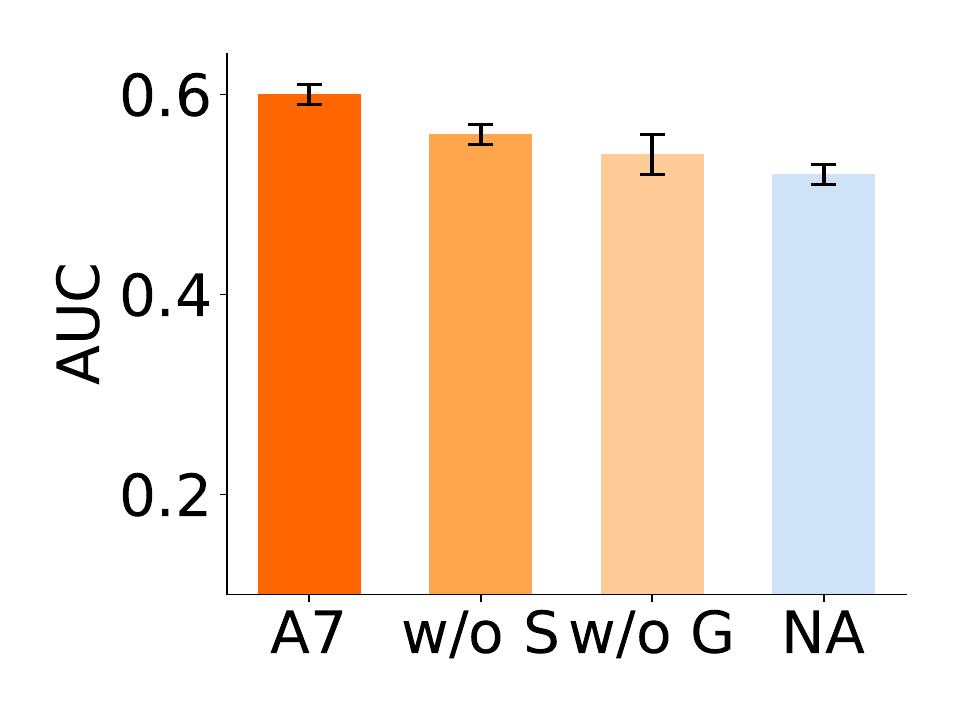}
    }%
    \caption{Ablation study on the contrastive advice selector~(S) and the  intrinsic reward generator~(G) for six Atari scenarios.}
    \label{fig:ablation}
\end{figure*}
\begin{figure*}[!t]
    \centering
    \hspace{-6mm}
    \subfigure[Freeway~(5k)]{
    \includegraphics[scale=0.26]{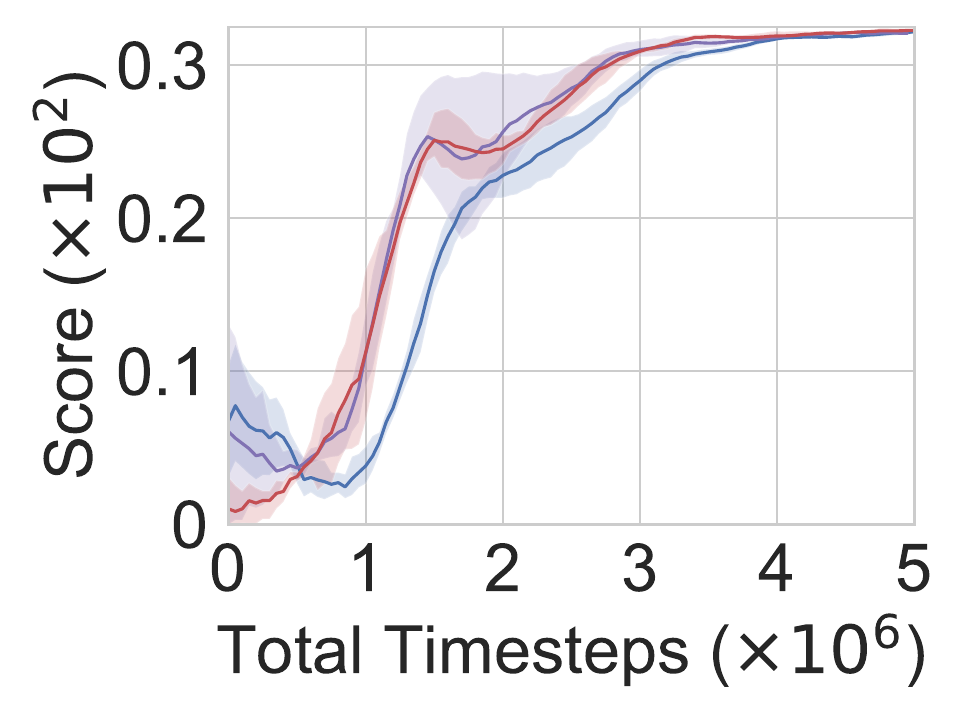}
    }
    \hspace{-3mm}
    \subfigure[Freeway~(15k)]{
    \includegraphics[scale=0.26]{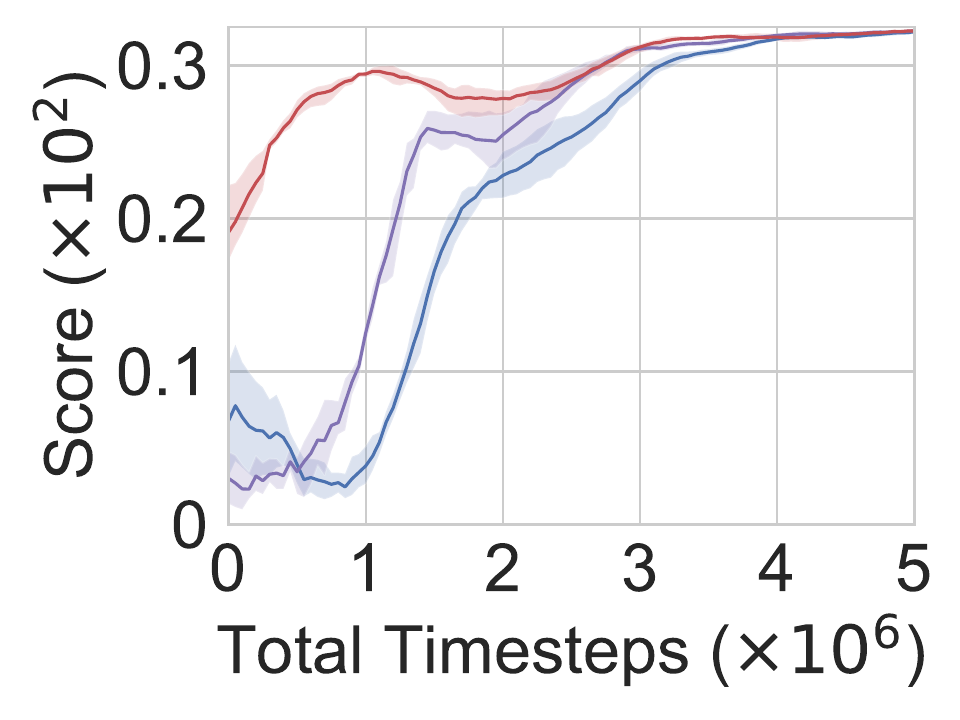}
    }
    \hspace{-3mm}
    \subfigure[Freeway~(25k)]{
    \includegraphics[scale=0.26]{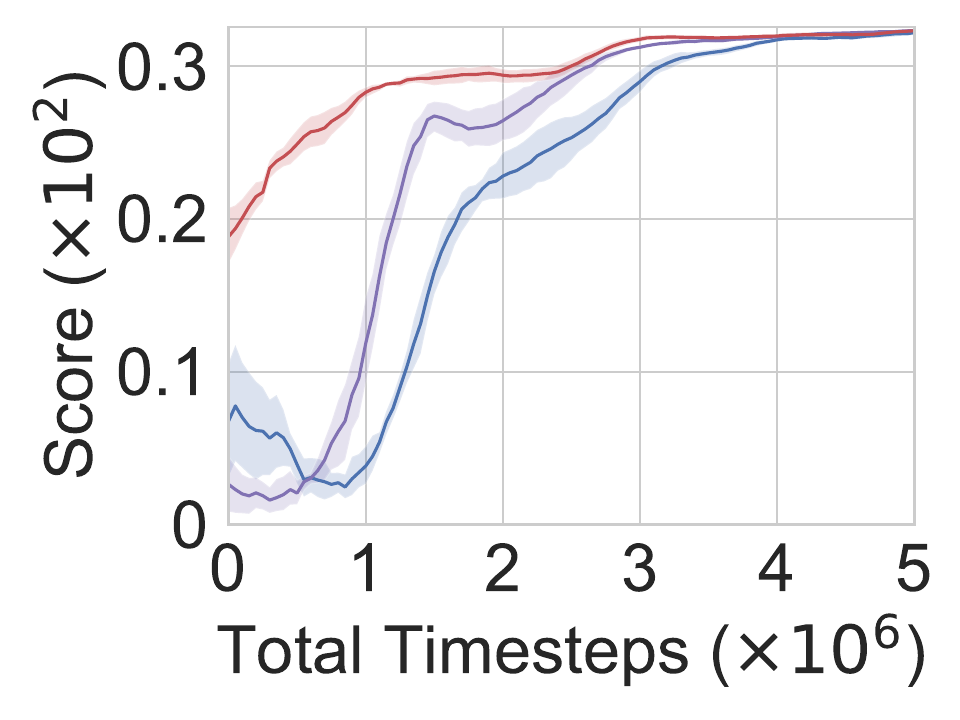}
    }
    \hspace{-3mm}
    \subfigure[Freeway~(35k)]{
    \includegraphics[scale=0.26]{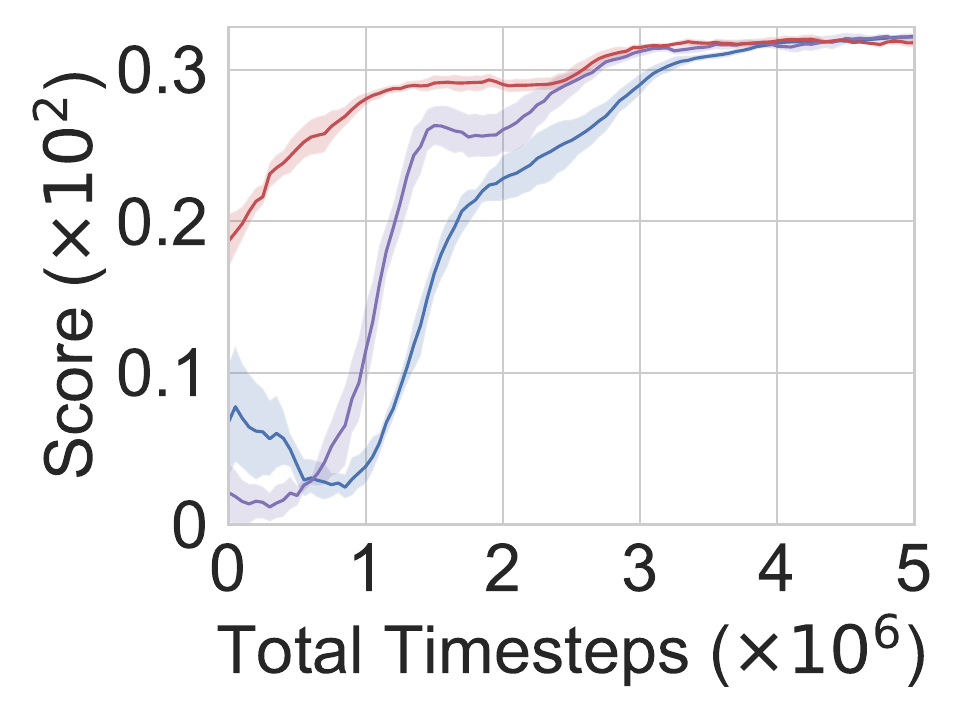}
    }

    \hspace{-3mm}
    \subfigure[Qbert~(5k)]{
    \includegraphics[scale=0.26]{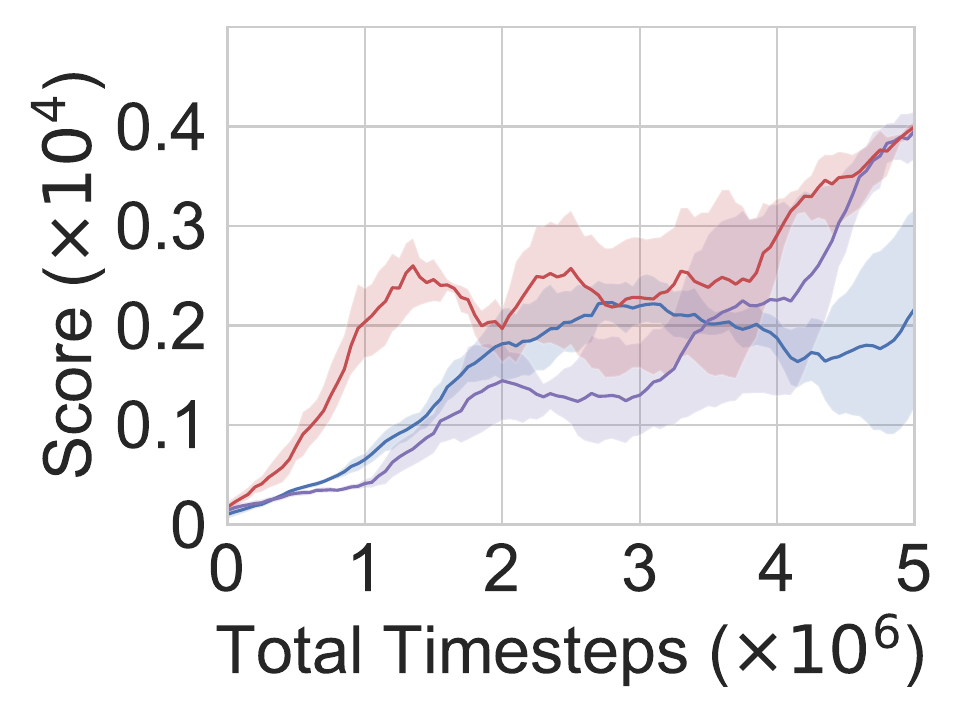}
    }
    \hspace{-3mm}
    \subfigure[Qbert~(15k)]{
    \includegraphics[scale=0.26]{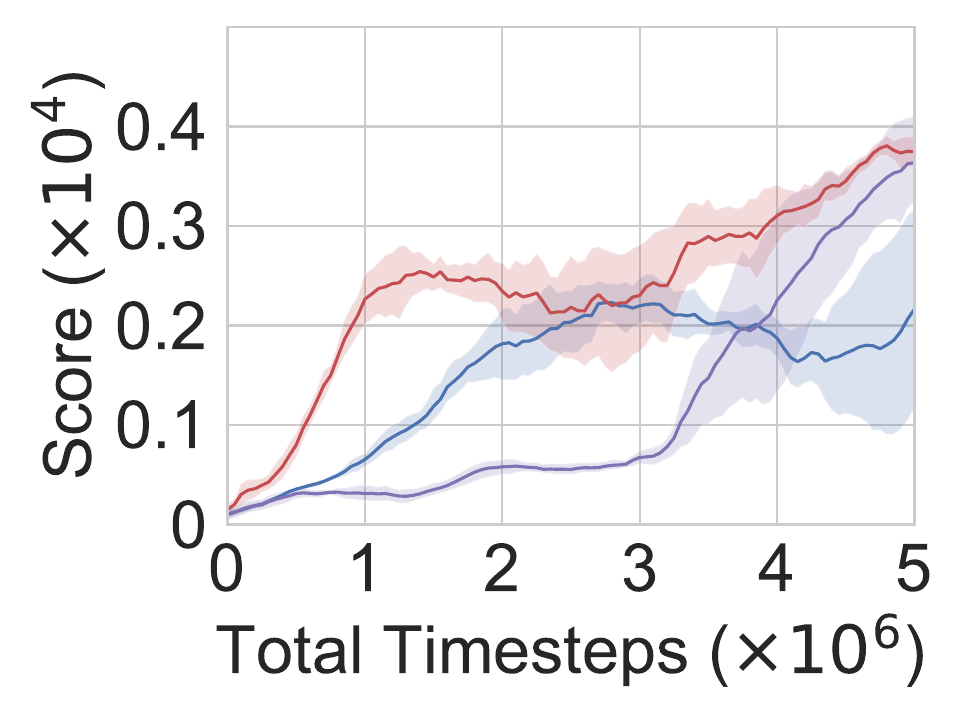}
    }
    \hspace{-3mm}
    \subfigure[Qbert~(25k)]{
    \includegraphics[scale=0.26]{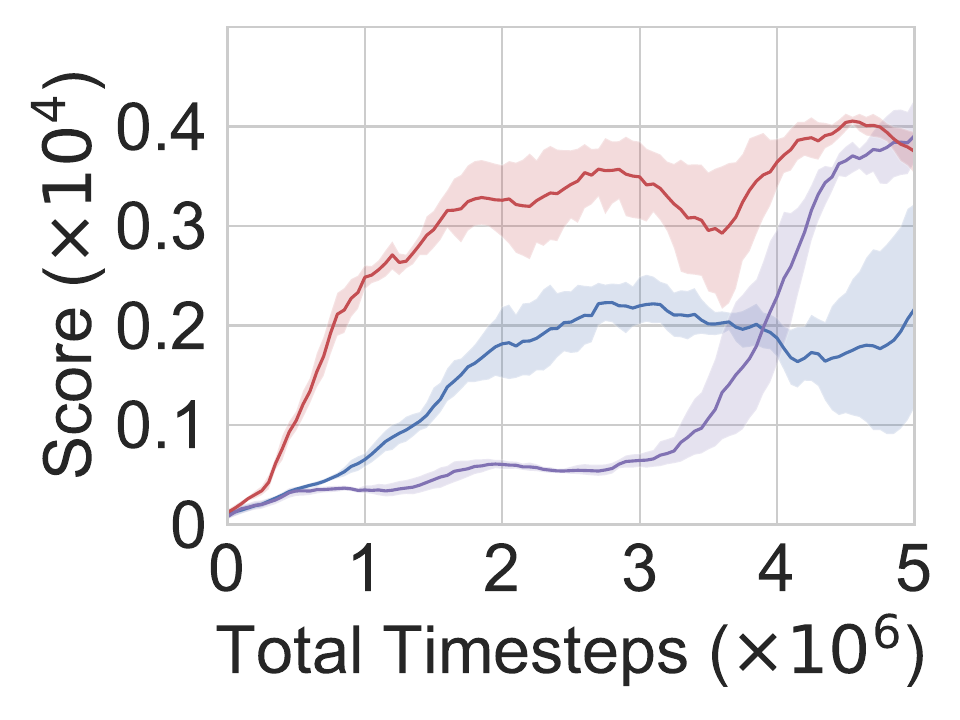}
    }
    \hspace{-3mm}
    \subfigure[Qbert~(35k)]{
    \includegraphics[scale=0.26]{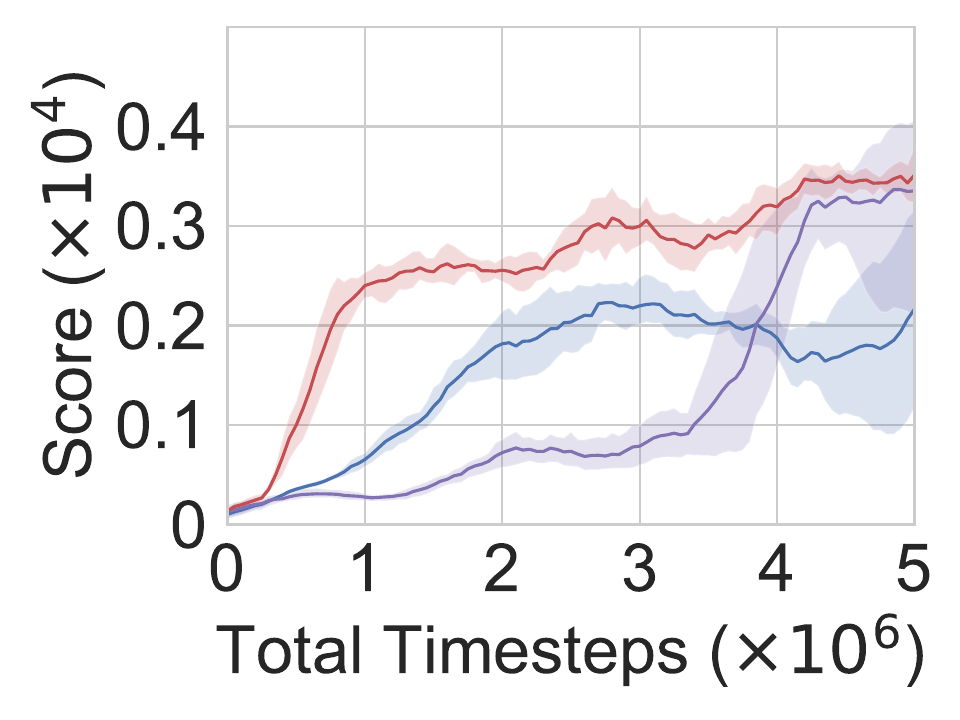}
    }
    \caption{The performance comparison of our proposed A7~(red line) , SUA-AIR~(purple line) and No Advising baseline~(blue line) under different advice budgets in the Freeway scenario and the Qbert scenario.}
    \label{fig:budget_ablation}
\end{figure*}
SUA-AIR exhibits the highest overall performance among the agent-specific methods. However, it demonstrates limited performance improvements when encountering specific environments like Enduro and SpaceInvaders.
This limitation arises from its inadequate coverage of the recommended state space.
In addition, we also listed the best evaluation scores achieved by all methods throughout the training phase across different scenarios in Table~\ref{tab:scores}. It is evident that, in all scenarios except for Pong, our method attained the highest scores, which also validates A7 improves the scores of the agent compared to existing methods.
To sum up, the experimental results suggest that our novel framework A7 amalgamates the advantages of agent-specific and agent-agnostic approaches, improving the sampling efficiency and accelerating the agent learning process to achieve non-trivial performance.

\subsection{Ablation Studies}
\paragraph{The contribution of different components}
To understand the superior performance of A7, we carry out ablation studies to test the contribution of its two main components: contrastive advice selector and intrinsic reward generator. The results are shown in Figure~\ref{fig:ablation}. By comparing A7 without contrastive advice selector~(replace the advice selection strategy of A7 with EA, and set the intrinsic reward to a fixed value) and without the intrinsic reward generator, we can conclude that neither of them alone can achieve the level of A7. This comparison highlights the effectiveness of our excellent advice selection strategy and the benefits obtained from combining it with the design of intrinsic rewards in achieving excellent results. Additionally, the AUC of both components exceeds that of NA, which demonstrates the effectiveness of both components in accelerating agent training and improving sampling efficiency.

\paragraph{The impact of different advice budgets} Moreover, to study the impact of different advice budgets on the performance of A7, we conduct an ablation study as shown in Figure~\ref{fig:budget_ablation}. The performance benefit of A7 experiences a substantial increase when the number of advice budgets rises from 5k to 25k. Conversely, the performance of SUA-AIR has shown minimal improvement. This can be attributed to the fact that the states selected by A7 for seeking advice can better represent the entire sample space.
However, the states selected by SUA-AIR are quite similar, resulting in a lack of improvement in terms of performance.
It is also noteworthy that the budget for A7 increased from 15k to 35k in Freeway and 25k to 35k in Qbert, but the consequent growth in performance was minimal.
This suggests that once a tipping point is reached, an increase in the number of advice results in a gradual decline in the growth of benefits.

\section{Conclusion}
In this work, we propose a novel framework called \textit{Agent-Aware trAining yet Agent-Agnostic Action Advising}~(A7) to alleviate the sampling inefficiency problem in DRL. A7 amalgamates the advantages of agent-specific and agent-agnostic methods, making it the first dedicated attempt to explicitly build the similarity of state features as the indicator for action advising. Specifically, the proposed A7 comprises two key components, namely, contrastive advice selector and intrinsic reward generator. The advice selector adopts contrastive learning to extract state features and then utilizes the feature similarity to determine whether to seek expert intervention at each step. The reward generator collects state-advice pairs chosen by the selector to train a reuse model. It then uses the model to provide advice again and further assigns intrinsic rewards for {advised} samples.
Experimental results on different environments show that A7 accelerates the training of agents more effectively and yields significantly high sampling efficiency compared with state-of-the-art competitors.

\textbf{Limitations and Future works.} The framework of our method needs training three models, which results in additional time consumption. The single inference time of A7 is 2.4ms, which is approximately 1.5 times that of the baseline with no advising. Additionally, action advising methods are limited to environments with discrete action spaces. Due to the expansive range of continuous actions, human teachers often struggle to provide precise continuous actions,  which can result in unfavorable outcomes when sub-optimal advice is given. Thus, an important future direction lies in the development of action advising methods designed for environments with continuous action spaces, broadening their applications in various practical domains.
Moreover, there is the potential for cost reduction if the teacher can only offer ambiguous advice, such as providing pair-wise preferences between different trajectories or giving several candidate actions. Learning from ambiguous advice presents a challenge for student agents.

\balance
\bibliographystyle{IEEEtranN}
\bibliography{a7}

\end{document}